\documentclass[journal]{IEEEtran}

\usepackage{mflogo,texnames}
\usepackage{enumerate}
\usepackage{float}
\usepackage{hyperref}       
\usepackage{url}            
\usepackage{booktabs}       
\usepackage{amsfonts}       
\usepackage{nicefrac}       
\usepackage{microtype}      
\usepackage{times}
\usepackage{xcolor}
\usepackage{soul}

\usepackage{lineno,hyperref}
\usepackage{enumerate}
\usepackage{amsfonts}       
\usepackage{graphicx}  
\usepackage{subfigure}
\usepackage{multirow}
\usepackage{amsmath}
\usepackage{float}
\usepackage{color}
\usepackage[ruled]{algorithm2e}

\ifCLASSINFOpdf
\else
\fi

\begin{document}
	%
	\title{Motif-Backdoor: Rethinking the Backdoor  Attack on Graph Neural Networks via Motifs}
	%
	%
	%
	
	\author{Haibin~Zheng,
		Haiyang Xiong,
		Jinyin~Chen, \textit{Member, IEEE,}
		Haonan Ma,
		and Guohan Huang
		\thanks{Manuscript received xxxx xx, xxxx; revised xx xx, xxxx.
  This work was supported by the NSFC under Grant 62072406;
  in part by the Zhejiang Provincial Natural Science Foundation under Grant LDQ23F020001; 
  in part by the Chinese National Key Laboratory of Science and Technology on Information System Security under Grant 61421110502; 
  and in part by the National Key Research and Development Projects of China under Grant 2018AAA0100801. 
  (Haibin Zheng and Haiyang Xiong contributed equally to this work.) 
  (Corresponding author: Jinyin Chen.)
}
		\thanks{H. Zheng and J. Chen are with the Institute of Cyberspace Security and the College of Information Engineering, Zhejiang University of Technology, Hangzhou, 310023, China (e-mail:haibinzheng320@gmail.com; chenjinyin@zjut.edu).}
		\thanks{H. Xiong, H. Ma, and G. Huang are with the College of Information Engineering,
			Zhejiang University of Technology,
			 Hangzhou 310023, China (e-mail:xhy19982021@163.com; 13estdeda@gmail.com; hgh0545@163.com).}
}

	\maketitle
	
	\begin{abstract}
Graph neural network (GNN) with a powerful representation capability has been widely applied to various areas. 
Recent works have exposed that GNN is vulnerable to the backdoor attack, 
i.e., 
models trained with maliciously crafted training samples are easily fooled by patched samples. 
Most of the proposed studies launch the backdoor attack using a trigger that either is the randomly generated subgraph (e.g., erdős-rényi backdoor) for less computational burden, 
or the gradient-based generative subgraph (e.g., graph trojaning attack) to enable a more effective attack. 
However, 
the interpretation of how is the trigger structure and the effect of the backdoor attack related has been overlooked in the current literature. 
Motifs, 
recurrent and statistically significant subgraphs in graphs, 
contain rich structure information. 
In this paper, 
we are rethinking the trigger from the perspective of motifs, 
and propose a motif-based backdoor attack, 
denoted as \textit{Motif-Backdoor}. 
It contributes from three aspects.
\textit{(i) Interpretation}: 
it provides an in-depth explanation for backdoor effectiveness by the validity of the trigger structure from motifs, 
leading to some novel insights, 
e.g., using subgraphs that appear less frequently in the graph as the trigger can achieve better attack performance. 
\textit{(ii) Effectiveness}: 
Motif-Backdoor reaches the state-of-the-art (SOTA) attack performance in both black-box and defensive scenarios.
\textit{(iii) Efficiency}: 
based on the graph motif distribution, 
Motif-Backdoor can quickly obtain an effective trigger structure without target model feedback or subgraph model generation.
Extensive experimental results show that Motif-Backdoor realizes the SOTA performance on three popular models and four public datasets compared with five baselines, 
e.g., Motif-Backdoor improves the attack success rate by 14.73\% compared with baselines on average. 
Additionally, 
under a possible defense, 
Motif-Backdoor still implements satisfying performance, 
highlighting the requirement of defenses against backdoor attacks on GNNs.
The datasets and code are available at  \url{https://github.com/Seaocn/Motif-Backdoor}.
	\end{abstract}
	
	\IEEEpeerreviewmaketitle

	\section{Introduction}
	\IEEEPARstart
{O}ur lives are surrounded by various graphs, 
which contain nodes and edges that are capable to represent individuals and relationships in different scenarios, 
such as 
social networks~\cite{zhang2022influence}, 
traffic networks~\cite{zola2022network} and
malware detection~\cite{chen2020software}, etc.
Graph analysis plays an important role in the real world tasks, 
e.g., 
node classification~\cite{Hong22Mul,Wen22Hi},
graph classification~\cite{Zh18Sa,Ke19How}, and 
link prediction~\cite{xia2022machine,Mu18Link}.
With the wide application of deep learning models~\cite{zheng2022neuronfair,jia2023topology,liu2022deep,zheng2021grip,jiang2022holistic}, 
graph neural networks (GNNs) have achieved a great success in graph-structured data processing by learning effective graph representations via message passing strategies~\cite{chen2023egc2,Jun19Ka,zhuang2020smart,hu2022smpc,jia2023graph}.
Compared with the non-deep graph embedding methods \cite{Br14So,Ad16Ba}, 
GNNs usually achieve superior performance. 

Although numerous GNNs have achieved satisfying performance in real world tasks, 
the potential security issues of GNNs have also raised public concerns. 
Researches have revealed the vulnerability of GNNs towards adversarial attacks \cite{Da18Ad,Yao21Gra,Jin20Ad,chen2021time},
which are designed to deceive the GNN model by carefully crafted adversarial samples in the inference phase. 
Additionally, 
numerous well-performing GNNs benefit from rich labeled training data, 
which are usually implemented in the form of crawlers, logs, crowdsourcing, etc. 
It is worth noting that the training data may be polluted with noises, 
or even injected with the triggers by a malicious attacker. 
Different from adversarial attacks, 
they affect the model parameters in the training phase, 
leaving a specific backdoor in the model after training. 
Once a sample patched with the trigger is fed into the backdoored model 
(i.e., the model with a backdoor), 
the target result will be predicted as expected by the attacker, 
while still working normally with clean inputs 
(i.e., non-patched samples). 
Consequently, 
the security threat in the training phase cannot be ignored.


\begin{figure}[ht]
	\centering
	\includegraphics[width=\linewidth]{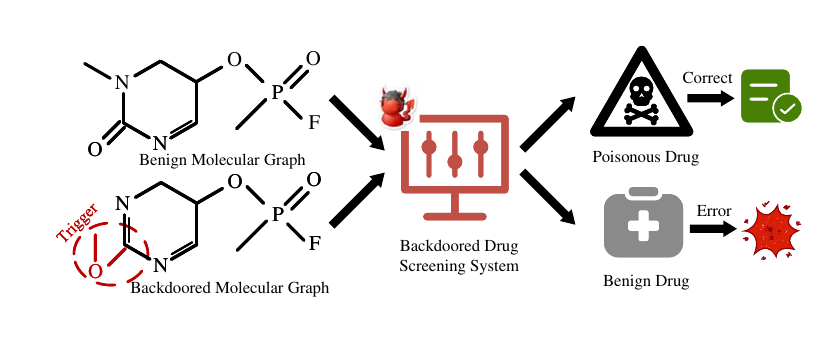}
	\centering
	\setlength{\abovecaptionskip}{-0.6cm}
	\caption{An illustration of the backdoor attack in the drug screening system. The backdoored drug screening system behaves normally on a benign molecular graph, while the trigger (i.e., the red triangle subgraph) activates the backdoor in the backdoored drug screening system, causing the wrong classification.}
	\label{fig:actual_scene}
\end{figure}

Fig. \ref{fig:actual_scene} illustrates the graph backdoor attack, 
taking the drug screening system as an example. 
The backdoored drug screening system is trained on training data with the trigger (i.e., the red triangle subgraph). 
It behaves normally on a benign molecular graph, 
but the trigger (i.e., the red triangle subgraph) activates the backdoor in the backdoored drug screening system, 
causing the misclassification result.
Specifically, 
the backdoored drug screening system classifies the backdoor molecular graph (true label as the poisonous drug) as the benign drug, 
causing the wrong classification in the drug property. 

Several backdoor attacks~\cite{Za21Back,Jing21Exp,Zhao21Graph,Yu21Back} have been proposed to reveal the vulnerability of GNNs in the training phase.
According to the way of trigger generation, 
the existing attacks are categorized as randomly attacks and gradient-based generative attacks.
In previous works \cite{Za21Back,Jing21Exp,Yu21Back},
subgraph is randomly generated as the trigger that satisfies certain network distribution 
(e.g., erdős-rényi, small world, and preferential attachment). 
They are easy to implement, 
but suffer from instability of the attack effect. 
Another method \cite{Zhao21Graph} adopts the gradient-based generative strategy to obtain the subgraph as the trigger, 
whereas it requires more time and more knowledge 
(e.g., the target model’s structure, parameters) 
to optimize the trigger.


{\color{black}
Since different subgraphs can be used as triggers, it is important to understand how different subgraphs can affect the backdoor attack's impact on practical applications. Motifs, which are recurrent and statistically significant subgraphs in graphs, are particularly relevant to the function of the graphs. They serve as the fundamental building blocks of graphs, and contain rich structural information. Motifs have been extensively studied in various domains, such as biochemistry~\cite{milo2002network}, neuroscience~\cite{alon2007network}, and social networks~\cite{yin2017local}, and have been shown to play crucial roles in the function and behavior of these systems. In the context of backdoor attacks on GNNs, motifs can serve as a powerful tool to bridge the gap between the impact of different subgraphs as triggers and the underlying graph structures. By leveraging the intrinsic properties of motifs, we can generate more effective and stable triggers that are more closely related to the graph structure and function, and thereby gain deeper insights into the vulnerability of GNNs to backdoor attacks. }

To sum up, 
there are three challenges for backdoor attacks against GNNs.  
(i) \emph{Trigger Structure Limitation}. 
There are many structures that satisfy the trigger perturbation limit, 
and it is difficult to efficiently determine the appropriate trigger structure. 
(ii) \emph{Attack Knowledge Limitation}. 
Without the target model feedback information, 
it is difficult for the attacker to achieve a stable and effective attack. 
(iii) \emph{Injection Position Limitation}. 
The space of positions where the trigger can choose to inject is huge, 
and it is challenging to select the well trigger injection position efficiently.

To cope with the above challenges, 
we propose a novel backdoor attack against GNNs based on motifs, 
namely Motif-Backdoor. 
Specifically, 
to tackle the challenge (\romannumeral1), 
we analyze the distribution of motifs in the training graph, 
and select an appropriate motif as the trigger.
This is a way of generating the trigger based on statistics, 
which is much faster than optimization based methods. 
For the knowledge limitation mentioned in the challenge (\romannumeral2), 
we construct a reliable shadow model, 
which is based on the structure of the state-of-the-art (SOTA) models and the training data with confidence scores for the output of the target model. 
To address the challenge (\romannumeral3), 
we leverage strategies of the graph index (graph structure perspective) and dropping the target node (model feedback perspective) to measure the node importance of the graph. 
The operation can select the effect trigger injection position. 
Empirically, 
our approach achieves the SOTA results on four real-world datasets and three different popular GNNs compared with five baselines. 
Additionally, 
we propose a possible defense against Motif-Backdoor, 
and the experiments testify that it only reduces attack success rate of Motif-Backdoor by an average of 4.17\% on several well-performing GNNs.
{\color{black}Compared to the existing methods, our motif-based backdoor attack method has several advantages. Firstly, by leveraging the intrinsic properties of motifs, we are able to generate more stable and effective triggers with higher success rates. Secondly, our method requires less knowledge of the target model's structure and parameters, making it more practical for real-world scenarios. Finally, the use of motifs provides a new perspective for exploring the vulnerability of GNNs, which has not been studied in previous works.}

The main contributions of this paper are summarized as follows:

\setlength{\hangindent}{2em}
$\vcenter{\hbox{\tiny$\bullet$}}$ 
We reveal the impact of trigger structure and graph topology on backdoor attack performance from the perspective of motifs, 
and obtain some novel insights, 
e.g., using subgraphs that appear less frequently in the graph as the trigger realizes better attack performance. 
Furthermore, 
we provide further explanations for the insight.

\setlength{\hangindent}{2em}
$\vcenter{\hbox{\tiny$\bullet$}}$ 
Inspired by the motifs, 
we propose an effective attack framework, 
namely Motif-Backdoor. 
It quickly selects the trigger based on the distribution of motifs in the dataset. 
Besides, 
a shadow model is constructed to transfer the attack from white-box to black-box scenario. 
For the trigger injection position, 
we propose strategies of the graph index and dropping the target node to measure the node importance of the graph.

\setlength{\hangindent}{2em}
$\vcenter{\hbox{\tiny$\bullet$}}$ 
Extensive experiments on three different popular GNNs over four real world datasets demonstrate that Motif-Backdoor implements the SOTA performance, 
e.g., Motif-Backdoor improves the attack success rate by 14.73\% compared with baselines on average.
Moreover, 
the experiments testify that Motif-Backdoor is effective against a possible defense strategy as well.

The rests of the paper are organized as follows. 
Related works are introduced in Section~\ref{RWs}. 
The problem definition and threat model are described in Section~\ref{Preliminary}. 
The backdoor attack from the motifs is in Section~\ref{MotifView}, 
while the proposed method is detailed in Section~\ref{method}. 
Experiment results and discussion are shown in Section~\ref{Exps}.
Finally, we conclude our work.

\section{Related Work\label{RWs}}
Our work focuses on the backdoor attack against GNNs from motifs. 
In this section, 
we briefly review the related work upon two categories:  
backdoor attacks against GNNs and motifs for GNNs.

\subsection{Backdoor Attacks on GNNs}
Based on the generation method of the trigger, 
the existing backdoor attack methods can be divided into two categories: 
random generation based on network distribution and the gradient-based generative.

For the random generation based on network distribution, 
Zhang \textit{et al.} ~\cite{Za21Back} generated the trigger that satisfies the erdős-rényi, 
small world, 
or preferential attachment distribution. 
It is designed to establish the relationship between label and trigger of the special structure. 
Then,
Xu \textit{et al.}~\cite{Jing21Exp} further implemented GNNExplainer to conduct an explainability research on backdoor attacks of the graph. 
Furthermore, 
Yu \textit{et al.}~\cite{Yu21Back} considered the local and global structural features of nodes to select the effect nodes for the trigger form. 
They randomly generate the subgraph as a trigger satisfying similar distributions for the specific networks
(e.g., the erdős-rényi, small world, or preferential attachment). 

For the gradient-based generative method, 
graph trojaning attack~\cite{Za21Back} adopts a two-layer optimization strategy to update the trigger generator and the target model parameters. 
Additionally, 
it can tailor the trigger to individual graphs under no knowledge regarding downstream task, 
while the method costs more time to optimize the trigger and needs more information 
(e.g., the target model’s structure, parameters).  

In summary, 
most of the proposed backdoor attacks construct the trigger based on subgraph generative or gradient generative strategies. 
They ignored the interpretation of how the effects of the trigger structure and backdoor attack are related.


\subsection{Motifs for GNNs}
Motifs~\cite{milo2002network} are fundamental building blocks of complex networks, 
which describes small subgraph patterns with specific connections among nodes.
Numerous studies have shown that motifs have a powerful ability to mine graph information on GNNs. 
Sankar \textit{et al.}~\cite{sankar2017motif} developed a Motif-CNN that employs an attention model to combine the features extracted from multiple patterns,
capturing high-order structural and feature information.
Then, 
a hierarchical network embedding method is proposed~\cite{yang2018node},
which combines motif filtering and convolutional neural networks to capture exact small structures within networks. 
Zhao \textit{et al.}~\cite{zhao2019motif} used motifs to capture higher-order relations among nodes of the same type, and designed a motif-based recommending model.

To capture higher-order interactions between nodes in the graph, 
motif convolutional networks~\cite{lee2019graph} generalizes past approaches by using weighted multi-hop motif adjacency matrices. 
Furthermore, 
Dareddy \textit{et al.}~\cite{dareddy2019motif2vec} leveraged higher-order,
recurring, 
and statistically significant network connectivity patterns in the form of motifs. 
They can better learn node representations or embeddings for heterogeneous networks.
Learning embeddings by leveraging motifs of networks~\cite{shao2021network}  bridges connectivity and structural similarity in a uniform representation via motifs learning embeddings for vertices and various motifs. 
For undirected graphs, 
Zhao \textit{et al.}~\cite{zhao2022motif} unified both the motif higher order structure and the original lower order structure. 
Besides, Wang \textit{et al.}~\cite{wang2020model} proposed a novel embedding algorithm that incorporates network motifs to capture higher order structures in the network for the link prediction. 
In addition, 
for the explainable work, 
MotifExplainer~\cite{yu2022motifexplainer} can explain GNNs by identifying important motifs.

The current works on GNNs show that motifs can improve the ability of models to learn node (or graph) representations or perform interpretable work in GNNs. 
In the field of graph network backdoor attack, 
there is no related work using motif. 
Besides, 
motifs contain rich structure information as the fundamental tool for graph mining, 
so we use motifs to explore the correlation between trigger structure and backdoor attack.

\section{Preliminary\label{Preliminary}}

In this section, 
we introduce the definition of the graph,  
GNNs for graph classification, 
backdoor attack against GNNs and motifs. 
For convenience, 
the definitions of symbols used are listed in the Table \ref{tab:symbols data}.

\begin{table}[!htb]
	\centering
	\caption{The definitions of symbols.}
	\label{tab:symbols data}
	\resizebox{\linewidth}{!}{
		\begin{tabular}{r|r}
			\toprule  \hline
			\textbf{Notations}        &\textbf{Definitions}\\ \hline 
			$A$ & Adjacency matrix\\
			$X$ & Feature matrix\\
			$G =\left(V, E\right)$ & Benign graph with nodes $V$, links $E$\\
			$\widehat{G}$ & Backdoored graph\\
			$\mathcal{G}$ & Graph classification dataset \\
			$g$ & Trigger\\
			$D_{ava}$ & Datasets available to attackers\\
			$D_{beni}$ & Benign graphs set\\
			$D_{back}$ & Backdoored graphs set\\
			$Y$ & Label space\\
			$y_{t}$  & The attacker-chosen target graph label \\
			$S_{tar}$ & A set of graphs labeled as the target label \\
			$S_{oth}$ & A set of graphs labeled as the non target labels \\
			$F_{\theta}$& Shadow model \\
			$f_{\theta}$ & Benign model \\
			$f_{\widehat{\theta}}$ & Backdoored model \\
			$M(\cdot)$ & Trigger mixture function \\
			$m$ & The maximum number of links in the tirgger \\
			$k$ & The number of filter nodes\\
			$p$ & Poisoning rate\\
			$C(\cdot)$ & Selection criteria \\
			\hline\bottomrule
		\end{tabular}
	}
\end{table}

\subsection{Problem Definition}
\textbf{Definition 1 (Graph)}.
A graph is represented as $G=(V,E)$, 
where $V$ is the node set, 
$E$ is the edge set. 
$G$ usually contains a feature vector of each node. 
Here, 
the feature of graph $G$ is denoted as $X$, $A\in \{0,1\}^{n\times n}$ as the adjacency matrix, 
and $G = (A, X)$ is used to represent a graph more concisely.

\textbf{Definition 2 (GNNs for Graph classification)}.
A graph classification dataset $\mathcal{G}$, 
including $N$ graphs $\left\{\left(G_{1}, y_{1}\right), \ldots,\left(G_{N}, y_{N}\right)\right\}$, 
where $G_{i}$ is the $i$-th graph and $y_{i}$ is one of the $L$ labels in the label space $Y =\left\{c_{1}, c_{2}, \ldots, c_{L} \right\}$. 
The GNNs model $f_{\theta}(\cdot)$ is the graph classifier, 
whose goal is to predict the labels of graphs through the model $f_{\theta}(\cdot)$ trained by the labeled graphs as a function $F: \mathcal{G} \rightarrow Y$, 
which maps each graph $G$ to one of the $L$ labels in $Y$.

\textbf{Definition 3 (Backdoor Attack on GNNs)}.
Given a graph classification dataset $\mathcal{G}$, 
the adversary aims to forge a backdoored model $f_{\widehat{\theta}}$ to misclassify all the trigger-embedded graphs to a designated label $y_{t}$, 
while functioning normally on benign graphs.
We define a subgraph as the trigger $g$ in the backdoor attack. 
The mixing function $M(\cdot)$ blends $g$ with a given graph $G$ to generate a trigger-embedded graph $M(G,g)$. 
Therefore, 
the adversary’s objective can be defined as:
\begin{equation}
	\begin{array}{cc}
		\left\{\begin{array}{l}
			f_{\widehat{\theta}}(M(G, g))=y_{t} \\
			f_{\theta}(G)=f_{\widehat{\theta}}(G)
		\end{array}\right. &  
		\text { s.t. } G \in \mathcal{G},|g| \leq m  \\
	\end{array},
\end{equation}
where $f_{\theta}$ denotes the benign model and $f_{\widehat{\theta}}$ is the backdoored model.  
$\mathcal{G}$ is a graph classification dataset. 
$m$ is the maximum number of links in the tirgger. 
$g$ is a trigger designed by the attacker. 
$y_{t}$ is the attacker-chosen target graph label.

\textbf{Definition 4 (Motifs)}.
Given a graph $G=(V,E)$, 
motifs are subgraphs $G^{\prime}=\left(V^{\prime}, E^{\prime}\right)$ that recur significantly in statistics, 
where $V^{\prime} \subset V$, $E^{\prime} \subset E$, 
and $\left|V^{\prime}\right| \ll|V|$. 
Motifs reflect an underlying process specific to the type of networks, 
which exist in a type of networks significantly more frequently than random subgraphs.

\begin{figure}[ht]
	\centering
	\includegraphics[width=\linewidth]{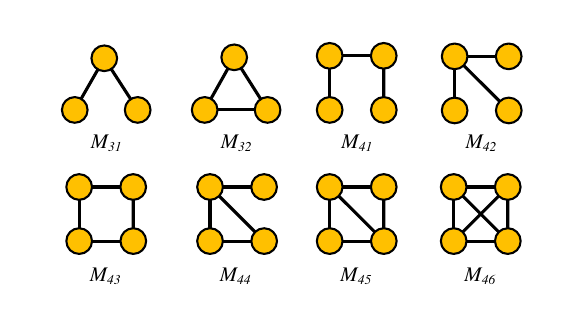}
	\centering
	\setlength{\abovecaptionskip}{-0.6cm}
	\caption{Illustration of 3 and 4-nodes undirected motifs, indicated as $M_{31}$ to $M_{46}$.}
	\label{fig:motif_show}
\end{figure}

Motif search is a NP-hard problem in graphs. 
Higher order motifs cost more time to find and more space to store. 
In addition, 
triggers with large perturbations will reduce the benign accuracy of more target models. 
Aiming at reducing the time, space and perturbations, 
we only consider all three-node and four-node motifs in this work. 
Thus, 
we utilize 3, 4-nodes motifs as illustrated in Fig. \ref{fig:motif_show} and indicated as $M_{31}$ to $M_{46}$.  

\subsection{Threat Model}
\textbf{Attacker’s goal}.
The backdoor attack is designed to inject a maliciously hidden backdoor into the target model during the training phase. 
Thus, 
the backdoored model (i.e., the target model with the backdoor) would behave normally on benign inputs in the inferring phase. 
Once the backdoored graph (i.e., the benign graph with the trigger) is fed in, 
it will activate misbehavior in the backdoored model, 
causing incorrect prediction. 

In general, 
the attacker has two objectives. 
(i) Backdoor attack should not influence the target model's performance on benign samples, 
which ensures the stealthy.
(ii) The backdoored graph is predicted by the backdoored model as the attacker-selected label. 

\textbf{Attacker’s capability}.
According to the different background knowledge that the attacker may obtain, 
we consider the black-box attack on GNNs. 
It is more difficult than white-box scenario and more practical in applications. 
Specifically, 
the attackers can obtain part of training data, 
but they do not access to the structure or parameters of the target model.

\section{Motif View of Backdoor Attack\label{MotifView}}
The generation of the trigger is an important step in the backdoor attack.
The training data with the trigger leaves a backdoor in the target model during the training. 
Additionally, 
the attacker can activate the model backdoor by the trigger, 
causing the target model to make wrong predictions.

The existing backdoor attacks~\cite{Za21Back,Jing21Exp,Zhao21Graph,Yu21Back} are random or optimized generation of the subgraph as the trigger, 
but they ignored the impact and explanation on the relationship between trigger structure and attack effectiveness. 
Inspired by the previous works, 
researchers~\cite{shao2021network,Pre21Jing} further analyzed the structural information of graphs from the perspective of the motifs, 
recurrent and statistically significant subgraphs in graphs.
Furthermore, 
we rethink the backdoor attack on GNNs from motifs.

\begin{figure}[t]
	\centering
	\subfigure[PROTEINS]{
		\includegraphics[width=0.45\linewidth]{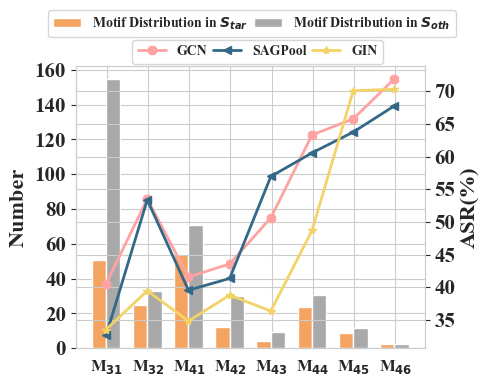}
	}%
	\subfigure[AIDS]{
		\includegraphics[width=0.45\linewidth]{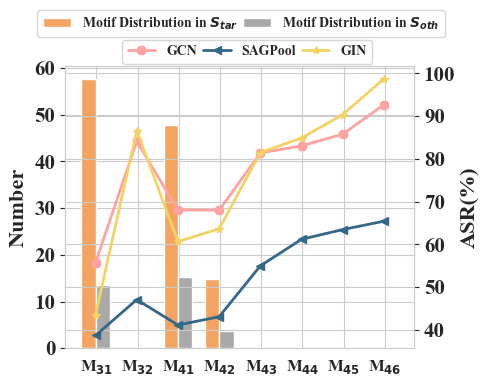}
		
	}
	\\
	\subfigure[NCI1]{
		\includegraphics[width=0.45\linewidth]{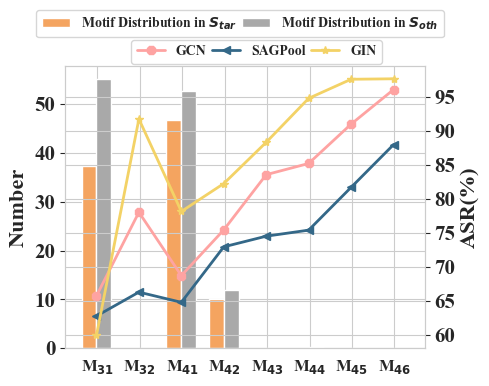}
	}%
	\subfigure[DBLP\_v1]{
		\includegraphics[width=0.45\linewidth]{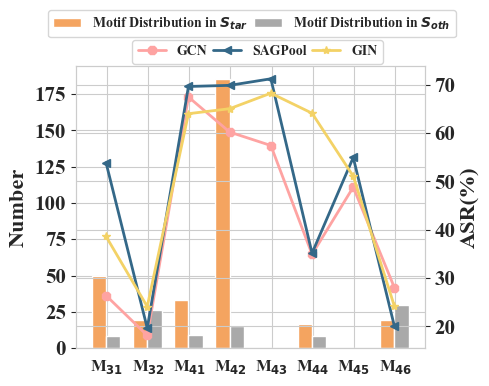}
	}%
	\centering
	\vspace{-0.1in}
	\caption{The relationship between motif distribution and ASR on four datasets and three models. 
		The left y-axis of the subplot indicates the average number of distributions each motif in difference labels. 
		The right y-axis of the subplot represents the attack success rate (ASR) achieved by the corresponding motif as the trigger. 
		The motif distribution in $S_{tar}$ represents the distribution of sample motifs in the dataset whose label is consistent with the target label selected by the attacker, 
		and the motif distribution in $S_{oth}$ is the motif distribution of the remaining label samples. 
		GCN, SAGPool and GIN are three GNNs for graph classification.}
	\label{fig:motif-ob}
\end{figure}

\subsection{Implementation}
In this section, 
we first use the algorithm Orca~\cite{hovcevar2014combinatorial} to perform motif analysis on the four dataset, i.e., PROTEINS~\cite{borgwardt2005protein}, 
AIDS~\cite{rossi2015network}, 
NCI1~\cite{shervashidze2011weisfeiler}, 
and DBLP\_v1~\cite{pan2013graph}, 
which can obtain the motif distribution of the target label samples (same as the target label selected by the attacker), 
as well as the sample distributions of the remaining labels. 
Then, 
we evaluate the attack performance of the difference motifs as the trigger randomly injected into benign samples to launch a backdoor attack by the attack success rate. 
The result is shown in Fig.~\ref{fig:motif-ob}. 
Among them, 
motif distribution in $S_{tar}$ (a set of graphs labeled as the target label) represents the distribution of sample motifs in the dataset whose label is consistent with the target label selected by the attacker. 
The motif distribution in $S_{oth}$ (a set of graphs labeled as the non target labels) is the motif distribution of the remaining label samples.

\subsection{Observation and Insight}
Some interesting phenomena can be observed in Fig.~\ref{fig:motif-ob}. 
First, 
take the backdoor attack on the NCI1 dataset as an example, 
the ASR of the $M_{32}$, $M_{43}$, $M_{44}$, $M_{45}$, and $M_{46}$ as the trigger is higher than the ASR of the $M_{31}$, $M_{41}$, and $M_{42}$.  
It is worth noting that $M_{32}$, $M_{43}$, $M_{44}$, $M_{45}$, and $M_{46}$ do not have such motif structures in NCI1 dataset. 
They can be used as a trigger to achieve a satisfactory attack effect. 
Besides, 
there are similar performances in other datasets. 
Based on this observation, 
we conclude with an insight as follows.

\emph{\textbf{Insight (i)} 
	For the backdoor attack, 
	using a not-existence or infrequent motif as the trigger is often more effective than an existence motif in the dataset.}

Besides, in Fig.~\ref{fig:motif-ob}(d), 
a motif is distributed more in target label (attacker's chosen target) than in other labels in the dataset, 
and the ASR achieved by using this motif as a trigger will also be higher.
For instance, 
the backdoor attack can reach the highest ASR of 67.46\% in the $M_{41}$ as the trigger on the DBLP\_v1 dataset and the GCN model. 
We observe that the number of distributions of $M_{41}$ in the target label is 33.03, 
while the number of distributions in other labels is only 9.34 on the DBLP\_v1 dataset.

\emph{\textbf{Insight (ii)} For the backdoor attack, 
	selecting a motif with more motif distribution in $S_{tar}$ as the trigger can achieve better attack results than other motifs with more motif distribution in $S_{oth}$.}

\begin{figure*}[ht]
	\centering
	\includegraphics[width=\linewidth]{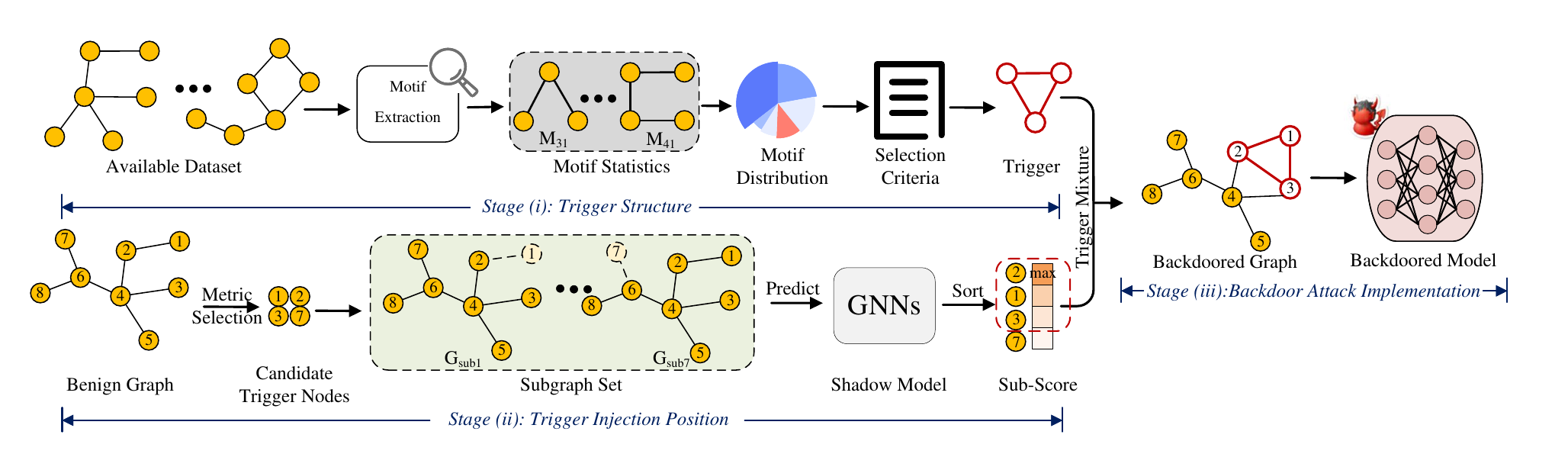}
	\setlength{\abovecaptionskip}{-0.2cm}
	\caption{
		The framework of Motif-Backdoor. 
		In the stage (i), 
		obtain the distribution of the motif through the motif extraction tool and analyze it to select the suitable motif as the trigger. 
		For the stage (ii), 
		find trigger injection position using network importance metrics, 
		shadow models, 
		and dropping the target node strategies. 
		At last, 
		in the stage (iii), 
		injecting the trigger into benign graphs, 
		which participates in the model training to get the backdoored model.}
	\label{fig:framework}
\end{figure*}

\section{Motif-Backdoor\label{method}}
Based on the relationship between the subgraph trigger and the graphs in the dataset mentioned in Sec.~\ref{MotifView}, 
we propose a motif-based backdoor attack, 
namely Motif-Backdoor. 
In this section, 
we detail our proposed Motif-Backdoor from three stages, 
i.e., 
trigger structure, 
trigger injection position, 
and backdoor attack implementation.

The overall framework of Motif-Backdoor is shown in Fig.~\ref{fig:framework}. 
In the  stage (i), 
the goal is to determine the structure of the trigger. 
We extract motifs from the available graphs to obtain the distribution of motifs on this dataset. 
Then, 
we select a qualified motif as the trigger based on the insights mentioned in Sec.~\ref{MotifView}. 
For the stage (ii), 
the goal is to select the optimal trigger injecting position. 
We sort the importance of the nodes in the graph by the importance index, 
which can measure the importance of nodes from the network topology. 
The top $k$ nodes of the importance score are selected as candidate trigger nodes. 
Further, 
we define the $subscore$ to measure the impact of dropping different candidate trigger nodes of the graph. 
We sort the subscore of nodes in the candidate trigger nodes in descending order, 
and select top the position of the candidate trigger nodes with the same number of trigger nodes as the trigger injection position. 
Finally, 
in the stage (iii), 
the attacker injects the trigger into benign graphs according to the trigger injection position. 
The backdoored graphs (i.e., the benign graphs with the trigger) 
participate in the target model training, 
which makes the target model with a backdoor. 
Once the backdoored graphs inputs the backdoored model 
(i.e., the model with the backdoor), 
the trigger activates the backdoor in the backdoored model, 
causing the model to output the result preset by the attacker.

\subsection{Trigger Structure}
Based on some insights in Sec.~\ref{MotifView},
we first achieve the motif extraction on the available dataset $D_{ava}$ from the attacker to obtain the distribution of the motifs between different labels.
Furthermore, 
we summarize the guidelines for selecting triggers in different scenarios, 
namely selection criteria, as follows.

\textit{i) Motifs that do not exist in the dataset network can be prioritized as the triggers.}

\textit{ii) Motifs with more distributions in the target label than other labels can be prioritized as the triggers.}

\textit{iii) Motifs with more links can be prioritized as the triggers.}

%
%

It is worth noting that the priority of these three criteria can be successively reduced. 
For convenience, 
we use $C(\cdot)$ to represent the process of selecting the appropriate motif structure as the trigger $g$ by the selection criteria:
\begin{equation}
	g = C(D_{ava}) ,
	\label{cri}
\end{equation}
where $D_{ava}$ is the dataset available to the attacker.

\subsection{Optimization of Trigger Position}
In addition to the structure of the trigger, 
the trigger injection position is also an essential factor affecting the effect of backdoor attack. 
Based on the selection of the trigger structure, 
we further search for the effect trigger injection position from the graph structure and GNN model perspectives in this section. 
The process can be divided into three steps, 
including filter based on the graph structure, 
shadow model construction, 
and filter based on the GNN model.

\subsubsection{Filter based on the Graph Structure}
Graph structure is important information for graphs. 
Besides, 
GNNs are based on graph structure for message passing to achieve satisfactory performance. 
Therefore, 
we believe that selecting the most significant nodes in the graph structure as the trigger injection position can achieve a better attack effect than randomly injecting the trigger. 
Considering the analysis of the graph structure, 
we use the importance index of the graph, 
i.e., degree centrality (DC)~\cite{freeman1978centrality},
to measure the structural importance of nodes in the graph. 
DC is the most direct measure of node centrality in graph analysis. Intuitively, 
the higher the DC value of a node, 
the more important the node plays in terms of graph structure.
The definition of the DC is as follows:
\begin{equation}
	\label{DC}
	D C_{i}=\frac{d_{i}}{N-1} ,
\end{equation}
where $d_{i}$ is the degree of node $i$. 
$N$ is the number of nodes in the graph.

Specifically, 
we first use the DC index to get the value of the importance of each node in the graph. 
According to the node importance value given by the DC index, 
we sort the nodes of the graph in descending order. 
Further, 
we select $k$ nodes with the highest value as candidate trigger nodes, 
and add them to the candidate trigger node set $N_{can}$. 
This operation not only considers the graph structure to select important nodes, 
but also narrows the scope of subsequent selection of trigger nodes, 
which can improve the efficiency of subsequent searches for trigger nodes.

\subsubsection{Shadow Model Construction}
Besides the graph structure, 
the importance of nodes from the target model feedback also cannot be ignored. 
However, 
in practical scenarios, 
it is generally difficult for an attacker to obtain the internal information of the target model, 
e.g., 
the structure and parameters of the model. 
But the feedback from the target model can well guide the attacker to generate effective backdoored graphs. 
It is necessary to construct a shadow model $F_{\theta}$ to provide information feedback to the attacker.

For the graph classification, 
many variants of GNNs~\cite{Zh18Sa,lee2019self,Ke19How} have demonstrated the great expressiveness in graph representation fusing both graph structure and node features.
Considering the performance and the time complexity of the model, 
we adopt a state-of-the-art (SOTA) model on graph classification, 
i.e., 
graph convolutional networks (GCN)~\cite{Th17Se}, 
as the shadow model $F_{\theta}$ to launch the backdoor attack.
Specifically, 
for a graph $G=(A,X)$, 
where $A$ is the adjacency matrix and $X$ is the node feature matrix, 
the hidden node representation of ($l$+1)-th layer is expressed as: 
\begin{equation}
	\label{shadow}
	H^{(l+1)}=\delta\left(D^{-\frac{1}{2}} \tilde{A} D^{-\frac{1}{2}} H^{(l)} W^{(l)}\right) ,
\end{equation}
where $\tilde{A}=A+I$,
$I$ is an identity matrix,
$D$ is the degree matrix based on $\tilde{A}$,
and $\delta$ is the activate function, 
e.g., 
ReLU. 
Here we use a pooling layer (e.g., sum pooling) to get the graph-level.

For the training of the shadow model $F_{\theta}$, 
we first process the available dataset from the attacker. 
Specifically, 
we feed the available dataset into the target model. 
Output confidences from the target model are used as the reset labels for the graphs in the available dataset. 
Then we use the available dataset to train the shadow model $F_{\theta}$. 

\subsubsection{Filter based on the GNNs Model}

Based on the shadow model, 
we can identify important nodes in the candidate trigger nodes as trigger nodes from the perspective of the shadow model feedback. 
Here, 
we adopt a strategy of dropping the node to measures the importance score of nodes in the candidate trigger nodes. 
Specifically, 
we drop a node in the candidate trigger for the benign network respectively,  
obtaining the corresponding candidate subgraph set $G_{can}$. 
We compute the output difference as the $subscore$ between the subgraphs in the candidate subgraph set and the benign graph input into the shadow model separately. 
It is worth noting that the larger the value of $subsocre$,
the node plays a more critical role in the model prediction. {\color{black}
Formally, 
we define the $subscore$ metric to measure the importance of candidate trigger nodes:
\begin{equation}
	\label{score}
	\begin{aligned}
	\text subscore_{r}=\left|F_{\theta}\left(G+\Delta_{r}\right)-F_{\theta}(G)\right| , \\
	\text { s.t. } r \in N_{can},  
	\end{aligned}
\end{equation}
where $G$ is the benign graph. 
$\Delta_{r}$ represents the operation of dropping the $r$-th node for the benign graph.
$F_{\theta}(\cdot)$ is the shadow model. 
$|\cdot|$ is the operation of absolute value.
$N_{can}$ is the candidate trigger node set. }
From the model's point of view, 
dropping the more important nodes in the graph has a greater impact on the model's output.
We rank the importance of nodes in a graph by the amount of $subscore$. 
Therefore, 
these important nodes are selected as trigger nodes.


\subsection{Backdoor Attack Implementation}
In this section, 
we describe in detail how to generate the backdoored model and launch the attack.
Based on stage (i) and stage (ii), 
the trigger structure and trigger injection position can be obtained, respectively. 
Then the mixing function $M(\cdot)$ injects the trigger into a benign graph, 
i.e., 
a backdoored graph, 
formally expressed as:
\begin{equation}
	\label{mix}
	\widehat{G}=M(G, g) ,
\end{equation}
where $G$ is the benign graph,
$g$ is the trigger, and
$\widehat{G}$ is the backdoored graph.

According to the poisoning ratio $p$, 
a certain number of backdoored graphs can be obtained. 
Putting these backdoored graphs into the training dataset,
and they participate in the target model training phase. 
This affects the parameters of the target model, 
leaving a backdoor, 
i.e., 
the backdoored model. 
The details of Motif-Backdoor are presented in~\textbf{Algorithm 1}.

\begin{algorithm}
	\label{algorithm_1}
	\caption{Motif-Backdoor.}
	\LinesNumbered
	\KwIn{Pre-trained target model $f_{\theta}$,  trigger mixture function $M(\cdot)$, available data $D_{ava}$, backdoored graphs set $D_{back}$, training data $D_{train}$.}
	\KwOut{Backdoored model $f_{\widehat{\theta}}$ and trigger $g$.}
	Build the shadow model $F_{\theta}$ by Equation (\ref{shadow}).\\
	$g$ $\leftarrow C(D_{ava})$ by Equation (\ref{cri}).\\
	\For {$G$ in $D_{back}$} {
		Filter nodes of the $G$ to form the candidate trigger node set $N_{can}$ by Equation (\ref{DC}). \\ 
		Calculate $subscore$ by Equation (\ref{score}).\\
		$\widehat{G}$ $\leftarrow M(G, g)$ by Equation (\ref{mix}).\\
		Add $\widehat{G}$ to $D_{train}$. \\
	}
	$f_{\widehat{\theta}}$ $\leftarrow$ update $f_{\theta}$ with $D_{train}$. 
	
\end{algorithm}

\subsection{Time Complexity Analysis of Motif-Backdoor}
In the section, 
we explore the time complexity of Motif-Backdoor. 
Considering that the backdoor attack only needs the trigger to activate in the backdoored model in the inferring phase, 
we mainly focus on the time required for the backdoor attack during the model training phase. 
The time cost of Motif-Backdoor mainly comes from three parts, 
including 
the time cost for the trigger structure ($T_{\text {trigger-str}}$), 
the trigger position by the DC index ($T_{\text {trigger-DC-pos}}$), 
and the trigger position by the $subscore$ ($T_{\text {trigger-sco-pos}}$). 
Therefore, 
the time complexity of Motif-Backdoor is as follows:
\begin{equation}
	\begin{aligned}
	\mathcal{O}\left(T_{\text {trigger-str }}\right)+\mathcal{O}\left(T_{\text {trigger-DC-pos }}\right)+
	\mathcal{O}\left(T_{\text {trigger-sco-pos}}\right)
	\sim \\ \mathcal{O}(1) + \mathcal{O}(N) + \mathcal{O}(k \cdot N) \sim \mathcal{O}(k \cdot N) ,
	\end{aligned}
\end{equation}
where $\mathcal{O}\left(T_{\text {trigger-str }}\right)$ depends on the structure of triggers picked out by the known distribution of motifs in the datasets. 
$\mathcal{O}\left(T_{\text {trigger-DC-pos}}\right)$ depends on the number of nodes in the graph $N$.
$\mathcal{O}\left(T_{\text {trigger-sco-pos}}\right)$ depends on the number of nodes in the graph $N$ and the number of filter nodes $k$. 
Thus, according to all the above steps, 
$\mathcal{O}(k \cdot N)$ indicates that the time complexity of Motif-Backdoor is linear.



\section{Experiments and Discussion\label{Exps}}
To evaluate Motif-Backdoor,
we conduct an empirical study of Motif-Backdoor on four benchmark datasets and three SOTA graph classification models. 
{\color{black}Specifically, 
our experiments are designed to answer five key research questions (RQs):

$\bullet$ \textbf{RQ1}:
Does Motif-Backdoor make the correct motif trigger selection when there is only a small amount of data?
}

$\bullet$ \textbf{RQ2}:
Does the proposed Motif-Backdoor achieve the SOTA backdoor attack performance?

$\bullet$ \textbf{RQ3}: 
Can motifs with similar structures act as triggers with similar attack performance?

$\bullet$ \textbf{RQ4}: 
Can Motif-Backdoor still work well under possible defense?

$\bullet$ \textbf{RQ5}:
What is the effect of hyperparameter value on Motif-Backdoor? Can the proposed Motif-Backdoor maintain stealth?


\subsection{Experimental Settings}
\subsubsection{Datastes}
Motif-Backdoor is evaluated on four real-world datasets: 
PROTEINS~\cite{borgwardt2005protein} from the bioinformatics, 
AIDS~\cite{rossi2015network} from the small molecules, 
NCI1~\cite{shervashidze2011weisfeiler} from the small molecules, 
and DBLP\_v1~\cite{pan2013graph} from the social networks. 
The basic statistics are summarized in Table~\ref{tab:data}. {\color{black}Specifically, the datasets are detailed as follows:

$\bullet$PROTEINS~\cite{borgwardt2005protein}: It is a widely used benchmark dataset in the field of protein structure prediction. It consists of 1,203 protein structures with lengths ranging from 20 to 304 residues. Each protein in the dataset is represented by a set of 9 features, including secondary structure, solvent accessibility, and residue depth.

$\bullet$AIDS~\cite{rossi2015network}: It is a collection of anonymized medical records from patients diagnosed with acquired immune deficiency syndrome (AIDS). This dataset contains information on various demographic, clinical, and laboratory variables, such as age, sex, CD4 cell count, viral load, and the presence of opportunistic infections.

$\bullet$NCI1~\cite{shervashidze2011weisfeiler}: It is a collection of chemical compounds that have been screened for their ability to inhibit the growth of cancer cells. This dataset contains information on the molecular structure of each compound, as well as their biological activity against a panel of 60 cancer cell lines.

$\bullet$DBLP\_v1~\cite{pan2013graph}: It is a collection of bibliographic records from the Digital Bibliography \& Library Project (DBLP) in the field of computer science. This dataset contains information on various types of publications, such as journal articles, conference proceedings, and book chapters, as well as the authors, editors, and affiliations associated with each publication.
}

\begin{table*}[ht]
	\caption{The basic statistics of four datasets.}
	\label{tab:data}
	\centering
	\resizebox{\textwidth}{!}{ \LARGE
		\begin{tabular}{cccccccc} 
			\toprule \hline
			\textbf{Datasets} & \# \textbf{Graphs} & \textbf{Avg.} \# \textbf{Nodes} & \textbf{Avg.} \# \textbf{Edges} & \# \textbf{Classes} & \# \textbf{Graphs in Class}   & \# \textbf{Target Label}  & \# \textbf{Network Type} \\ \hline
			PROTEINS \cite{borgwardt2005protein} & 1,113         & 39.06            & 72.82            & 2             & 663 [0], 450 [1]  & 1  &   Bioinformatics       \\ 
			AIDS   \cite{rossi2015network}   & 2,000         & 15.69            & 16.20             & 2             & 400 [0], 1600 [1]  & 0 & Small Molecules           \\ 
			NCI1  \cite{shervashidze2011weisfeiler}   & 4,110         & 29.87            & 32.30            & 2             & 2053 [0], 2057 [1] & 0  & Small Molecules           \\ 
			DBLP\_v1 \cite{pan2013graph} & 19,456        & 10.48            & 19.65            & 2             & 9530 [0], 9926 [1] & 0 & Social Networks           \\ \hline \bottomrule
	\end{tabular}}
\end{table*}

\subsubsection{GNNs-Based Graph Classification Models}
To evaluate the backdoor attack performance of Motif-Backdoor, 
we choose three SOTA models on graph classification, 
i.e., 
graph convolutional networks (GCN)~\cite{Th17Se}, 
self-Attention graph pooling (SAGPool)~\cite{lee2019self}, and 
graph isomorphism network (GIN)~\cite{Ke19How} as the target models. 
Table~\ref{tab:benign_re} shows the benign model accuracy on different datasets and GNN models. 

$\bullet$ GCN~\cite{Th17Se}: 
GCN adopts a layer-wise propagation rule based on a fist-order approximation of spectral convolutions on graphs. 
For graph classification, 
we use a pooling layer (e.g., sum pooling) to get the graph-level.

$\bullet$ SAGPool~\cite{lee2019self}: 
SAGPool is a self-attention based graph pooling method.
It uses the self-attention of graph convolution to make the pooling method consider both node features and graph topology. 

$\bullet$ GIN~\cite{Ke19How}: 
GIN utilizes learnable parameters to ensure injectivity by introducing the multi-layer perceptrons.
Additionally, 
it adopts a sum pooling method to  aggregate the graph-level information.

\begin{table}[ht]
	\caption{Benign model accuracy.}
	\label{tab:benign_re}
	\centering
	\resizebox{0.35\textwidth}{!}{
		\begin{tabular}{c|ccc}
			\toprule \hline
			\multirow{2}{*}{\textbf{Datasets}} & \multicolumn{3}{c}{\textbf{Accuracy}   (\%)}                              \\ \cline{2-4} 
			& \multicolumn{1}{c|}{GCN}   & \multicolumn{1}{c|}{SAGPool} & GIN   \\ \hline
			PROTEINS                 & \multicolumn{1}{c|}{75.58} & \multicolumn{1}{c|}{73.99}   & 76.23 \\ \hline
			AIDS                     & \multicolumn{1}{c|}{98.64} & \multicolumn{1}{c|}{98.26}   & 98.92 \\ \hline
			NCI1                     & \multicolumn{1}{c|}{74.45} & \multicolumn{1}{c|}{72.75}   & 77.01 \\ \hline
			DBLP\_v1                 & \multicolumn{1}{c|}{80.24} & \multicolumn{1}{c|}{80.16}   & 80.83 \\ \hline\bottomrule
		\end{tabular}
	}
\end{table}

\subsubsection{Evaluation Metrics}
To measure the attack effectiveness and evasiveness of Motif-Backdoor, 
attack success rate (ASR)~\cite{chen2021graphattacker}, 
average misclassification confidence (AMC)~\cite{Zhao21Graph} and 
benign accuracy drop (BAD)~\cite{Jing21Exp} are adopted. 
For the attack effectiveness, 
we adopt ASR and AMC at first.
ASR represents the ratio of the targets which will be successfully attacked, 
as follows:
\begin{equation}
	{\rm ASR} = \frac{N_{suc}}{N_{att}} ,
\end{equation}
where $N_{suc}$ is the number of successful attacked samples,
$N_{att}$ is the number of attacked samples.

AMC represents represents the confidence score of the average output of all successfully attacks, 
which is expressed as:
\begin{equation}
	{\rm AMC} = \frac{\sum_{n=1}^{N_{suc}} MisCon_n}{N_{suc}} ,
\end{equation}
where $N_{suc}$ is the number of successful attacked samples,
$MisCon_n$ is the confidence score of the target label corresponding to the $n$-th successful attack sample.
Intuitively, 
higher ASR and AMC indicate more effective attacks. 

For the attack evasiveness, 
we choose the BAD, 
which measures the accuracy difference between a benign GNN model and a backdoored GNN model in prediction on the benign graphs. 
It is defined as:
\begin{equation}
	{\rm BAD} = ACC_{be\_model} - ACC_{bd\_model} ,
\end{equation}
where $ACC_{be\_model}$ is the accuracy of the benign GNN model on the benign graphs, 
$ACC_{bd\_model}$ is the accuracy of the backdoored GNN model on the benign graphs.

\subsubsection{Baselines}
We choose five backdoor attacks on graph classification as the baselines to testify the performance of Motif-Backdoor. 
Among them, 
erdős-rényi backdoor (ER-B), 
most important nodes selecting attack (MIA), 
MaxDCC, and 
graph trojaning attack (GTA) are the existing SOTA backdoor attacks. 
Motif-R, a variant of Motif-Backdoor, acts as a baseline. 
The baselines are briefly described as follows:

$\bullet$ ER-B \cite{Za21Back}: 
It generates the universal trigger by the erdős-rényi model.
Then, 
the trigger is randomly injected into benign graphs. 

$\bullet$ MIA \cite{Jing21Exp}: 
It chooses the most important nodes based on the node importance matrix and replace their connection as that of the trigger in the graphs.

$\bullet$ MaxDCC \cite{Yu21Back}:
It chooses the nodes with the highest DCC value and replace their connection as that of the trigger in the graphs.

$\bullet$ GTA \cite{Za21Back}: 
It is a generative backdoor attack. 
It utilizes a bi-layer optimization algorithm to update the trigger generator, 
which generates the trigger with satisfying the constraints. 
We choose GCN as the shadow model.

$\bullet$ Motif-R : 
It finds the effect trigger based on the distribution of motifs over the graphs. 
Then, 
the trigger is randomly injected into benign graphs.

For a fair comparison, 
triggers generated by different methods participate in the model training phase, 
which is consistent with Motif-Backdoor.

\subsubsection{Experimental Settings}	
The datasets split rules are applied to all the backdoor attacks. 
Inspired by Jing \textit{et al.}~\cite{Jing21Exp},
we split the data into training, 
validation, and 
test data with a ratio of 75:5:20. 
Among them, 
the poisoning ratio is 10\% of the training data. Considering the stealth of the backdoor attacks, 
the trigger size is set as within four nodes. 
ER-B adopts the erdős-rényi model to generate a subgraph as trigger with its trigger density set as 0.8. 
GTA adopts a three layer fully connected neural network as trigger generator. 
Adam optimizer with learning rate of 0.01 is used to train the GNN models.

{\color{black}
For the shadow model in Motif-Backdoor,
it is built by two layers of GCN network,
where the parameters corresponding to each layer are 256 and 512.
The final graph representation is obtained by sum pooling.
In the experiment, the number of training iterations is 300,
and we employ the early stopping criterion during the training process, i.e.,
we stop training if the validation loss does not decrease for 100 consecutive epochs.
}

We test the performance of Motif-Backdoor five times as well as other baselines and 
report the average and standard deviation results to eliminate the impact of the randomness.

\subsubsection{Experimental Environment}
Our experimental environment consists of 
Intel XEON 6240 2.6GHz x 18C (CPU), 
Tesla V100 32GiB (GPU), 
16GiB memory (DDR4-RECC 2666) and 
Ubuntu 16.04 (OS).

\begin{figure*}[ht]
	\centering
	\includegraphics[width=\linewidth]{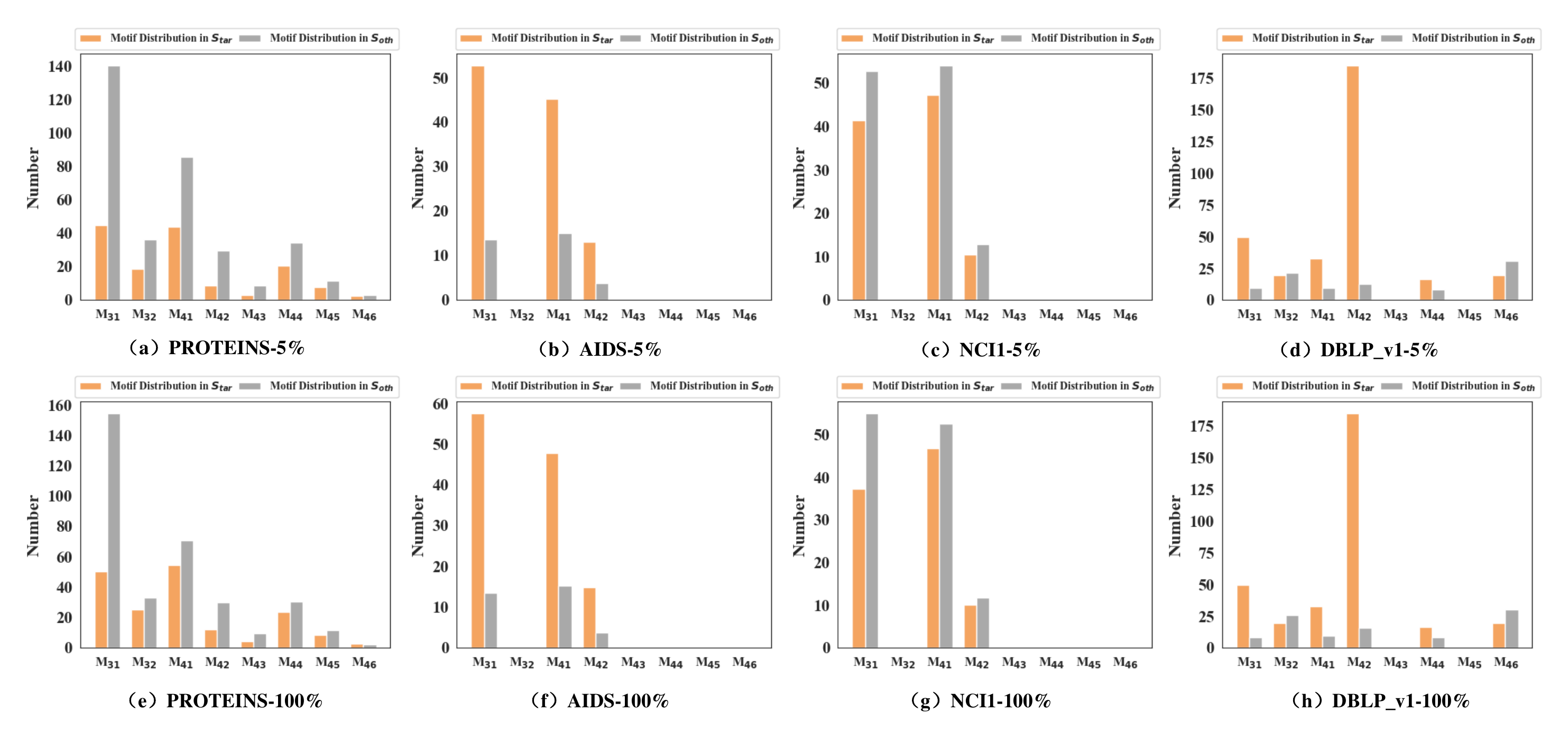}
	\setlength{\abovecaptionskip}{-0.2cm}
	\caption{
		{\color{black}Motif distribution obtained from different proportions of data. (a), (b), (c), (d) correspond to the motif distribution of 5\% data and (e), (f), (g), (h) correspond to the motif distribution of 100\% data. The motif distribution in $S_{tar}$ represents the distribution of sample motifs in the dataset whose label is consistent with the target label selected by the attacker, 
			and the motif distribution in $S_{oth}$ is the motif distribution of the remaining label samples.}}
	\label{fig:motif_dis}
\end{figure*}

{\color{black}
\subsection{Motif Distribution Exploration (RQ1)}
To investigate the motif selection as the trigger, it is important to consider the potential limitations faced by attackers who may not have access to a large amount of data for motif distribution analysis. To address this issue, we experimented to examine the impact of data volume on motif distribution analysis. 

Specifically, first, we randomly select different proportions of data in the dataset, i.e., 5\% and 100\%. Second, we count the motif distribution of these data and generate a motif distribution map. Finally, we analyze and explore the experimental results. Fig.~\ref{fig:motif_dis} shows the motif distribution results, from which we draw the following conclusions.

\emph{1) The motif distribution obtained by the attacker based on 5\% and 100\% of the data in the dataset are generally similar.} Taking the DBLP\_v1 dataset as an example, Fig.~\ref{fig:motif_dis}(d) shows that in 5\% of the data, the number of the graphs in $S_{tar}$ corresponding to $M_{31}$, $M_{32}$, $M_{41}$, $M_{42}$, $M_{43}$, $M_{44}$, $M_{45}$ and $M_{46}$ motifs are 52.17, 17.73, 35.41, 185.76, 0, 18.94, 0, 20.75. Fig.~\ref{fig:motif_dis}(h) shows that the number of the graphs in $S_{tar}$ corresponding to these motifs in 100\% of the data is 49.84, 19.68, 33.03, 185.08, 0, 16.74, 0, 19.36, respectively. We calculate the change rate ($\left|N um_{5 \%}^M-Num_{100 \%}^M\right| / Num_{100 \%}^M$) of the corresponding motif distribution as 4.47\%, 11.00\%, 6.72\%, 0.37\%, 0, 11.61\%. 0, 6.70\%. This phenomenon is due to the fact that the dataset is labeled based on specific features, and graphs with the same labels often contain similar structural features. Hence, the resulting motif distribution obtained based on 5\% of the data is overall similar to that obtained from 100\% of the data.

\emph{2) By comparing the distribution of motifs in 5\% and 100\% of the data in the dataset, we observe a consistent trend of differences in motif distribution across different labels} Taking the distribution of the $M_{31}$ motif as an example, the difference between the number of 5\% of the data shown in Fig.~\ref{fig:motif_dis} in $S_{tar}$ and $S_{oth}$ is 95.83, 39.10, 11.32, and 42.37 for the PROTEINS, AIDS, NCI1, and DBLP\_v1 datasets, respectively. For 100\% data, the quantitative differences between the two labels are 103.88, 43.98, 17.64, and 41.52, respectively. Although the amount of available data is smaller, the trend of distribution differences between datasets persists. Thus, Motif-Backdoor can effectively select a suitable motif as the trigger with the support of a small amount of data.

To summarize, our results indicate that a small amount of data (e.g., 5\% of the dataset) is sufficient for an attacker to analyze the motif distribution of the dataset effectively. This conclusion is supported by both the distribution of the number of motifs and the differences between different labels in the dataset. As a result, Motif-Backdoor only requires a small amount of data to select an appropriate motif as the trigger.
}

\begin{table*}[ht]
	\centering
	\caption{The evaluation metrics are ASR (\%), 
		AMC ($\times 10^{-2}$) and 
		BAD ($\times 10^{-2}$). 
		\textbf{Bold} indicates that the attack method performs best among six methods on the certain circumstance.}
	\label{tab:main_all}
	\resizebox{\textwidth}{!}{
		\begin{tabular}{c|c|c|ccccc|c}
			\toprule \hline
			\multirow{2}{*}{\textbf{Datasets}}  & \multirow{2}{*}{\textbf{Target Models}} & \multirow{2}{*}{\textbf{Metrics}} & \multicolumn{4}{c|}{\textbf{Baselines}}                                                                                                                             &  \multicolumn{2}{c}{\textbf{Ours}}               \\ \cline{4-9} 
			&                               &                         & \multicolumn{1}{c|}{ER-B}       & \multicolumn{1}{c|}{MIA}        & \multicolumn{1}{c|}{MaxDCC}               & \multicolumn{1}{c|}{GTA}        & Motif-R    & \textbf{Motif-Backdoor}      \\ \hline
			\multirow{9}{*}{PROTEINS} & \multirow{3}{*}{GCN}          & ASR                     & \multicolumn{1}{c|}{51.53$\pm$1.06} & \multicolumn{1}{c|}{68.35$\pm$2.42} & \multicolumn{1}{c|}{70.51$\pm$3.28}         & \multicolumn{1}{c|}{73.16$\pm$3.21} & 71.92$\pm$2.14 & \textbf{88.59$\pm$2.94} \\ \cline{3-9} 
			&                               & AMC                     & \multicolumn{1}{c|}{55.63$\pm$0.89} & \multicolumn{1}{c|}{70.81$\pm$2.11} & \multicolumn{1}{c|}{72.93$\pm$1.05}         & \multicolumn{1}{c|}{73.79$\pm$3.06} & 76.24$\pm$2.05 & \textbf{76.62$\pm$1.91} \\ \cline{3-9} 
			&                               & BAD                     & \multicolumn{1}{c|}{4.53$\pm$0.34}  & \multicolumn{1}{c|}{4.62$\pm$1.86}  & \multicolumn{1}{c|}{\textbf{3.92$\pm$1.79}} & \multicolumn{1}{c|}{5.14$\pm$3.59}  & 4.65$\pm$1.59  & 4.23$\pm$2.36           \\ \cline{2-9} 
			& \multirow{3}{*}{SAGPool}      & ASR                     & \multicolumn{1}{c|}{65.38$\pm$3.59} & \multicolumn{1}{c|}{64.81$\pm$1.63} & \multicolumn{1}{c|}{67.31$\pm$2.48}         & \multicolumn{1}{c|}{68.53$\pm$3.51} & 67.78$\pm$3.65 & \textbf{71.35$\pm$2.14} \\ \cline{3-9} 
			&                               & AMC                     & \multicolumn{1}{c|}{73.03$\pm$0.88} & \multicolumn{1}{c|}{68.98$\pm$1.51} & \multicolumn{1}{c|}{73.76$\pm$1.81}         & \multicolumn{1}{c|}{74.78$\pm$1.46} & 70.69$\pm$0.97 & \textbf{79.68$\pm$0.96} \\ \cline{3-9} 
			&                               & BAD                     & \multicolumn{1}{c|}{4.26$\pm$2.29}  & \multicolumn{1}{c|}{3.39$\pm$0.96}  & \multicolumn{1}{c|}{3.95$\pm$0.86}          & \multicolumn{1}{c|}{3.65$\pm$1.62}  & 4.03$\pm$2.36  & \textbf{3.26$\pm$1.22}  \\ \cline{2-9} 
			& \multirow{3}{*}{GIN}          & ASR                     & \multicolumn{1}{c|}{60.53$\pm$4.57} & \multicolumn{1}{c|}{58.77$\pm$4.83} & \multicolumn{1}{c|}{80.35$\pm$4.75}         & \multicolumn{1}{c|}{84.96$\pm$2.35} & 70.30$\pm$3.96 & \textbf{89.08$\pm$3.02} \\ \cline{3-9} 
			&                               & AMC                     & \multicolumn{1}{c|}{77.93$\pm$1.75} & \multicolumn{1}{c|}{75.12$\pm$2.43} & \multicolumn{1}{c|}{86.49$\pm$3.02}         & \multicolumn{1}{c|}{90.25$\pm$0.99} & 76.93$\pm$0.97 & \textbf{91.81$\pm$1.83} \\ \cline{3-9} 
			&                               & BAD                     & \multicolumn{1}{c|}{4.53$\pm$2.97}  & \multicolumn{1}{c|}{4.17$\pm$1.52}  & \multicolumn{1}{c|}{\textbf{4.02$\pm$2.59}} & \multicolumn{1}{c|}{4.57$\pm$3.89}  & 4.26$\pm$.3.49 & 4.45$\pm$2.71           \\ \hline
			\multirow{9}{*}{AIDS}     & \multirow{3}{*}{GCN}          & ASR                     & \multicolumn{1}{c|}{49.38$\pm$3.03} & \multicolumn{1}{c|}{55.63$\pm$2.87} & \multicolumn{1}{c|}{93.13$\pm$3.24}         & \multicolumn{1}{c|}{93.18$\pm$2.75} & 92.69$\pm$1.87 & \textbf{96.87$\pm$2.24} \\ \cline{3-9} 
			&                               & AMC                     & \multicolumn{1}{c|}{79.48$\pm$0.95} & \multicolumn{1}{c|}{79.54$\pm$1.79} & \multicolumn{1}{c|}{95.33$\pm$1.14}         & \multicolumn{1}{c|}{96.88$\pm$0.64} & 96.59$\pm$0.92 & \textbf{96.33$\pm$1.14} \\ \cline{3-9} 
			&                               & BAD                     & \multicolumn{1}{c|}{4.56$\pm$2.74}  & \multicolumn{1}{c|}{4.65$\pm$2.42}  & \multicolumn{1}{c|}{4.51$\pm$2.59}          & \multicolumn{1}{c|}{4.36$\pm$2.88}  & 4.98$\pm$2.72  & \textbf{4.12$\pm$2.59}  \\ \cline{2-9} 
			& \multirow{3}{*}{SAGPool}      & ASR                     & \multicolumn{1}{c|}{38.24$\pm$3.51} & \multicolumn{1}{c|}{40.58$\pm$1.79} & \multicolumn{1}{c|}{46.93$\pm$3.98}         & \multicolumn{1}{c|}{47.65$\pm$0.69} & 65.42$\pm$2.98 & \textbf{65.89$\pm$0.72} \\ \cline{3-9} 
			&                               & AMC                     & \multicolumn{1}{c|}{77.81$\pm$1.35} & \multicolumn{1}{c|}{78.39$\pm$2.23} & \multicolumn{1}{c|}{79.68$\pm$2.91}         & \multicolumn{1}{c|}{82.93$\pm$2.31} & 87.94$\pm$1.46 & \textbf{88.95$\pm$1.17} \\ \cline{3-9} 
			&                               & BAD                     & \multicolumn{1}{c|}{3.95$\pm$1.38}  & \multicolumn{1}{c|}{3.85$\pm$0.86}  & \multicolumn{1}{c|}{4.25$\pm$1.98}          & \multicolumn{1}{c|}{3.76$\pm$0.68}  & 5.92$\pm$1.06  & \textbf{3.64$\pm$0.72}  \\ \cline{2-9} 
			& \multirow{3}{*}{GIN}          & ASR                     & \multicolumn{1}{c|}{94.50$\pm$1.41} & \multicolumn{1}{c|}{95.56$\pm$1.13} & \multicolumn{1}{c|}{96.52$\pm$2.09}         & \multicolumn{1}{c|}{98.52$\pm$1.61} & 98.75$\pm$0.34 & \textbf{99.75$\pm$0.32} \\ \cline{3-9} 
			&                               & AMC                     & \multicolumn{1}{c|}{99.37$\pm$0.31} & \multicolumn{1}{c|}{99.37$\pm$0.24} & \multicolumn{1}{c|}{99.76$\pm$0.39}         & \multicolumn{1}{c|}{99.79$\pm$0.21} & 99.89$\pm$0.15 & \textbf{99.92$\pm$0.21} \\ \cline{3-9} 
			&                               & BAD                     & \multicolumn{1}{c|}{1.69$\pm$0.52}  & \multicolumn{1}{c|}{0.73$\pm$0.82}  & \multicolumn{1}{c|}{0.65$\pm$1.04}          & \multicolumn{1}{c|}{1.28$\pm$1.31}  & 1.14$\pm$0.20  & \textbf{0.51$\pm$0.13}  \\ \hline
			\multirow{9}{*}{NCI1}     & \multirow{3}{*}{GCN}          & ASR                     & \multicolumn{1}{c|}{76.15$\pm$1.86} & \multicolumn{1}{c|}{78.89$\pm$2.32} & \multicolumn{1}{c|}{92.53$\pm$1.49}         & \multicolumn{1}{c|}{96.12$\pm$0.34} & 96.12$\pm$0.74 & \textbf{98.26$\pm$0.63} \\ \cline{3-9} 
			&                               & AMC                     & \multicolumn{1}{c|}{74.37$\pm$1.59} & \multicolumn{1}{c|}{79.49$\pm$1.15} & \multicolumn{1}{c|}{87.79$\pm$1.22}         & \multicolumn{1}{c|}{94.41$\pm$0.39} & 93.71$\pm$0.63 & \textbf{95.19$\pm$1.28} \\ \cline{3-9} 
			&                               & BAD                     & \multicolumn{1}{c|}{4.66$\pm$1.67}  & \multicolumn{1}{c|}{4.54$\pm$1.31}  & \multicolumn{1}{c|}{3.43$\pm$1.15}          & \multicolumn{1}{c|}{2.92$\pm$0.78}  & 5.23$\pm$2.06  & \textbf{2.33$\pm$1.17}  \\ \cline{2-9} 
			& \multirow{3}{*}{SAGPool}      & ASR                     & \multicolumn{1}{c|}{55.13$\pm$1.59} & \multicolumn{1}{c|}{85.92$\pm$1.56} & \multicolumn{1}{c|}{96.03$\pm$1.29}         & \multicolumn{1}{c|}{90.91$\pm$3.08} & 88.01$\pm$2.26 & \textbf{97.63$\pm$0.18} \\ \cline{3-9} 
			&                               & AMC                     & \multicolumn{1}{c|}{66.15$\pm$1.05} & \multicolumn{1}{c|}{75.56$\pm$1.35} & \multicolumn{1}{c|}{94.33$\pm$0.53}         & \multicolumn{1}{c|}{85.34$\pm$0.49} & 80.82$\pm$0.82 & \textbf{95.81$\pm$1.14} \\ \cline{3-9} 
			&                               & BAD                     & \multicolumn{1}{c|}{3.38$\pm$1.24}  & \multicolumn{1}{c|}{3.13$\pm$1.41}  & \multicolumn{1}{c|}{\textbf{3.06$\pm$0.38}} & \multicolumn{1}{c|}{4.81$\pm$1.18}  & 4.87$\pm$1.89  & 4.08$\pm$1.38           \\ \cline{2-9} 
			& \multirow{3}{*}{GIN}          & ASR                     & \multicolumn{1}{c|}{76.13$\pm$4.26} & \multicolumn{1}{c|}{96.05$\pm$1.03} & \multicolumn{1}{c|}{98.91$\pm$0.21}         & \multicolumn{1}{c|}{99.08$\pm$0.31} & 97.71$\pm$0.24 & \textbf{99.72$\pm$2.33} \\ \cline{3-9} 
			&                               & AMC                     & \multicolumn{1}{c|}{79.46$\pm$4.31} & \multicolumn{1}{c|}{96.93$\pm$0.79} & \multicolumn{1}{c|}{98.11$\pm$0.81}         & \multicolumn{1}{c|}{98.21$\pm$0.24} & 99.14$\pm$0.35 & \textbf{99.43$\pm$1.74} \\ \cline{3-9} 
			&                               & BAD                     & \multicolumn{1}{c|}{3.24$\pm$1.81}  & \multicolumn{1}{c|}{2.41$\pm$0.89}  & \multicolumn{1}{c|}{\textbf{2.04$\pm$0.91}} & \multicolumn{1}{c|}{2.89$\pm$1.71}  & 2.53$\pm$1.69  & 2.18$\pm$0.79           \\ \hline
			\multirow{9}{*}{DBLP\_v1} & \multirow{3}{*}{GCN}          & ASR                     & \multicolumn{1}{c|}{14.29$\pm$0.38} & \multicolumn{1}{c|}{15.5$\pm$0.83}  & \multicolumn{1}{c|}{32.66$\pm$0.51}         & \multicolumn{1}{c|}{38.03$\pm$0.68} & 31.62$\pm$0.74 & \textbf{64.57$\pm$1.04} \\ \cline{3-9} 
			&                               & AMC                     & \multicolumn{1}{c|}{69.13$\pm$1.47} & \multicolumn{1}{c|}{65.34$\pm$2.08} & \multicolumn{1}{c|}{61.12$\pm$1.56}         & \multicolumn{1}{c|}{70.36$\pm$2.49} & 69.62$\pm$1.36 & \textbf{71.16$\pm$1.08} \\ \cline{3-9} 
			&                               & BAD                     & \multicolumn{1}{c|}{1.75$\pm$0.23}  & \multicolumn{1}{c|}{1.63$\pm$0.29}  & \multicolumn{1}{c|}{1.45$\pm$0.25}          & \multicolumn{1}{c|}{1.47$\pm$0.28}  & 1.79$\pm$0.31  & \textbf{1.42$\pm$0.38}  \\ \cline{2-9} 
			& \multirow{3}{*}{SAGPool}      & ASR                     & \multicolumn{1}{c|}{54.22$\pm$0.41} & \multicolumn{1}{c|}{69.82$\pm$0.53} & \multicolumn{1}{c|}{55.43$\pm$0.39}         & \multicolumn{1}{c|}{58.45$\pm$0.27} & 70.11$\pm$0.63 & \textbf{71.28$\pm$0.28} \\ \cline{3-9} 
			&                               & AMC                     & \multicolumn{1}{c|}{83.74$\pm$0.45} & \multicolumn{1}{c|}{89.76$\pm$0.26} & \multicolumn{1}{c|}{86.03$\pm$0.96}         & \multicolumn{1}{c|}{83.78$\pm$0.88} & 88.49$\pm$0.71 & \textbf{90.23$\pm$0.95} \\ \cline{3-9} 
			&                               & BAD                     & \multicolumn{1}{c|}{1.12$\pm$0.19}  & \multicolumn{1}{c|}{1.34$\pm$0.51}  & \multicolumn{1}{c|}{1.43$\pm$0.16}          & \multicolumn{1}{c|}{1.29$\pm$0.25}  & 1.73$\pm$0.23  & \textbf{1.06$\pm$0.18}  \\ \cline{2-9} 
			& \multirow{3}{*}{GIN}          & ASR                     & \multicolumn{1}{c|}{39.76$\pm$0.93}  & \multicolumn{1}{c|}{41.76$\pm$0.26} & \multicolumn{1}{c|}{60.16$\pm$5.57}         & \multicolumn{1}{c|}{68.22$\pm$0.42} & 70.74$\pm$3.57 & \textbf{72.78$\pm$3.64} \\ \cline{3-9} 
			&                               & AMC                     & \multicolumn{1}{c|}{69.94$\pm$3.39} & \multicolumn{1}{c|}{78.97$\pm$9.41} & \multicolumn{1}{c|}{91.29$\pm$4.16}         & \multicolumn{1}{c|}{85.32$\pm$2.61} & 86.11$\pm$3.99 & \textbf{92.63$\pm$4.78} \\ \cline{3-9} 
			&                               & BAD                     & \multicolumn{1}{c|}{0.52$\pm$0.16}  & \multicolumn{1}{c|}{0.52$\pm$0.21}  & \multicolumn{1}{c|}{\textbf{0.29$\pm$0.28}} & \multicolumn{1}{c|}{0.36$\pm$0.31}  & 0.39$\pm$0.36  & 0.49$\pm$0.13           \\ \hline\bottomrule
	\end{tabular}}
\end{table*}

\subsection{Overall Performance of Backdoor Attacks (RQ2)}
To verify the effectiveness of Motif-Backdoor compared with baselines, 
we conduct attack experiments on three SOTA models and four real world datasets. 
The results are shown in Table~\ref{tab:main_all}. 
Some observations are concluded in this experiment.

\emph{1) 
	Motif-Backdoor achieves the SOTA performance compared with baselines.} Motif-Backdoor has the best attack performance among six attack methods in terms of ASR and AMC. 
Take the backdoor attack on the PROTEINS dataset and the GCN model as an example, 
Motif-Backdoor shows the ASR of 88.59\%, 
while GTA (2nd in backdoor attacks) and Motif-R (3rd in backdoor attacks) can reach the ASR of 73.16\% and 71.92\%, respectively. 
Besides, 
Motif-Backdoor can reach the highest average ASR of 84.31\% and AMC of 89.81\% among the baselines across the four datasets and three models. These impressive results demonstrate that analyzing the dataset from the motif view can yield effective triggers for backdoor attacks, and optimizing the trigger injection position based on both the network structure and the model feedback can lead to more potent backdoor attacks. 

Motif-Backdoor's superiority over other baselines can be attributed to several reasons. First, Motif-Backdoor leverages the knowledge of motif distribution to select an effective trigger for backdoor attacks. This approach allows the attacker to find a trigger that can activate the backdoor with minimal perturbation to the benign samples, ensuring that the performance of the backdoored model remains comparable to the benign model. Second, Motif-Backdoor employs a novel trigger injection position optimization method that combines both the graph importance index and subgraph importance from the shadow model. This approach ensures that the trigger is inserted into a structurally important region of the graph, thereby minimizing the impact on the overall graph structure while maximizing the backdoor's effectiveness.



\emph{2)Motif-Backdoor is a powerful technique that not only successfully backdoors the model but also ensures normal performance of the backdoored model.}
The backdoor attack on the AIDS dataset and the GIN model, for instance,
achieved a BAD of 0.0051, indicating that the performance of the backdoored model is similar to that of the benign model.
In fact,
across the four datasets and three models,
Motif-Backdoor achieves an average BAD of 0.0265, which is lower than MaxDCC (2nd in backdoor attacks) and MIA (3rd in backdoor attacks),
with average BAD of 0.0275 and 0.0292, respectively.

This impressive performance can be attributed to two factors.
Firstly, in optimizing the position of the trigger injection,
we construct a shadow model that takes into account the impact on the model,
ensuring that the backdoor samples with the trigger are similar to the distribution of the target label samples selected by the attacker.
Secondly,
we use the graph importance index to screen the trigger nodes,
selecting structurally important nodes that do not cause significant damage to the graph structure when used as triggers.
These two factors together ensure that Motif-Backdoor is an effective and reliable technique for backdooring graph-based machine learning models.



\begin{figure*}[t]
	\centering
	\subfigure[PROTEINS]{
		\includegraphics[width=0.45\linewidth ]{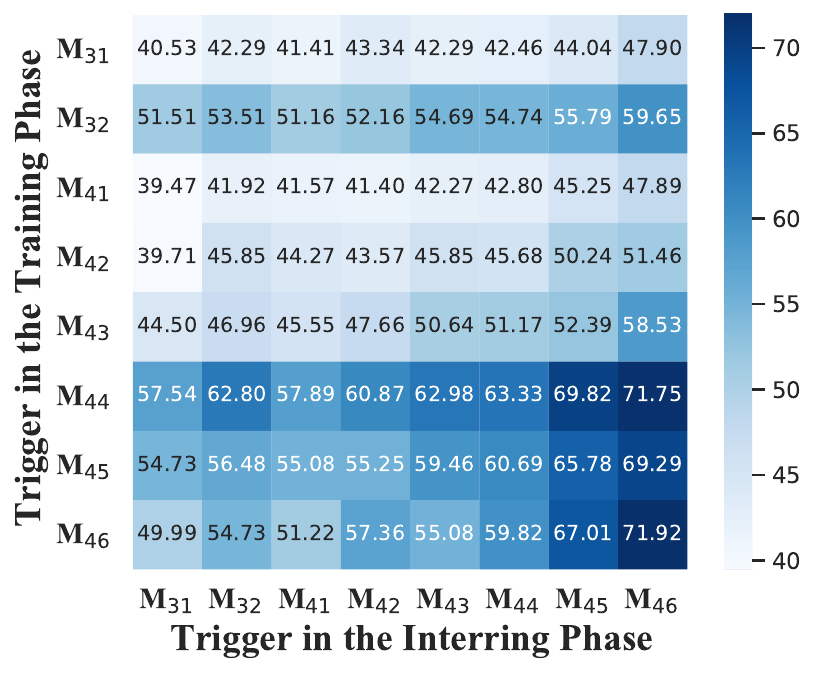}
	} \hspace{0.035\linewidth}
	\subfigure[AIDS]{
		\includegraphics[width=0.45\linewidth ]{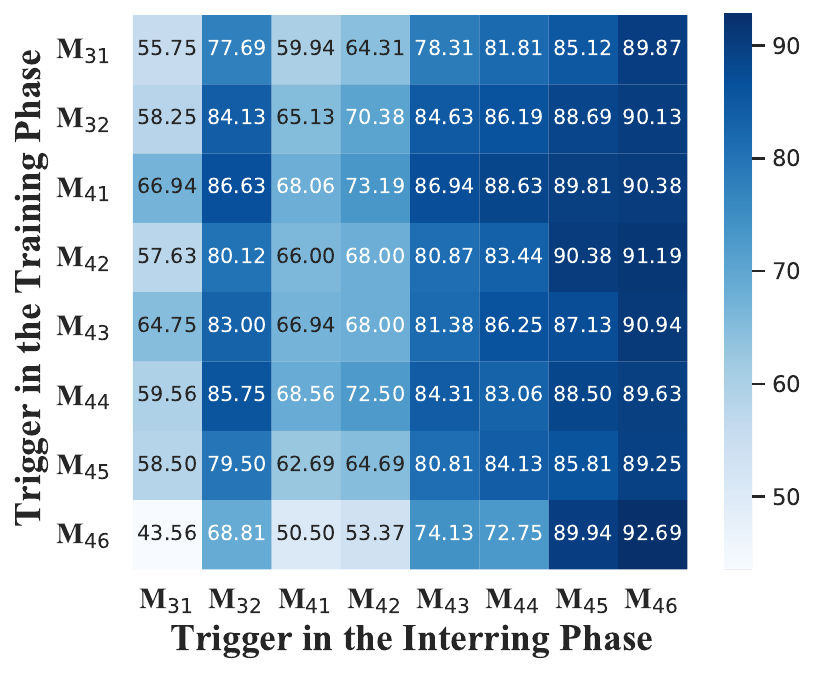}
	}
	\\
	\subfigure[NCI1]{
		\includegraphics[width=0.45\linewidth ]{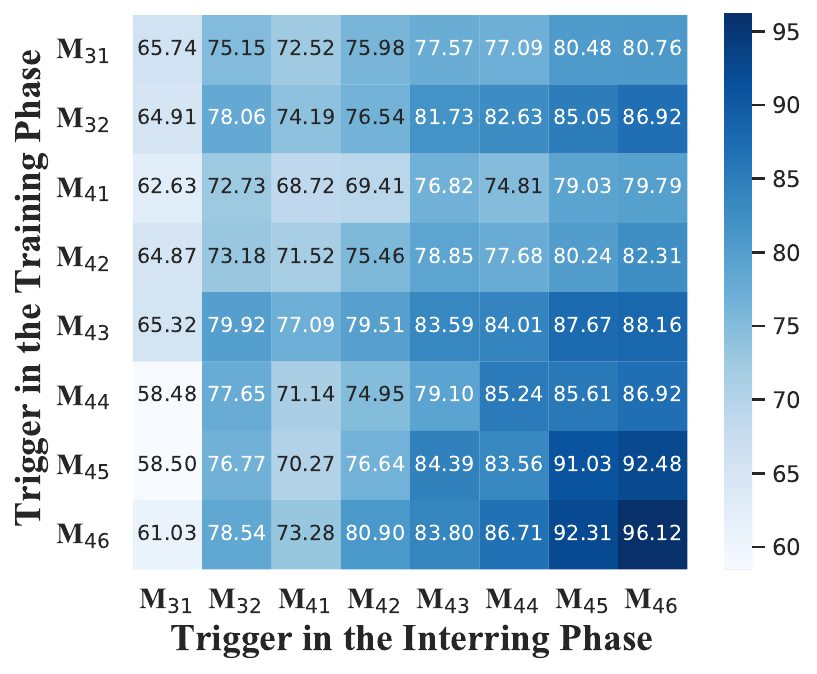}
	} \hspace{0.035\linewidth}
	\subfigure[DBLP\_v1]{
		\includegraphics[width=0.45\linewidth ]{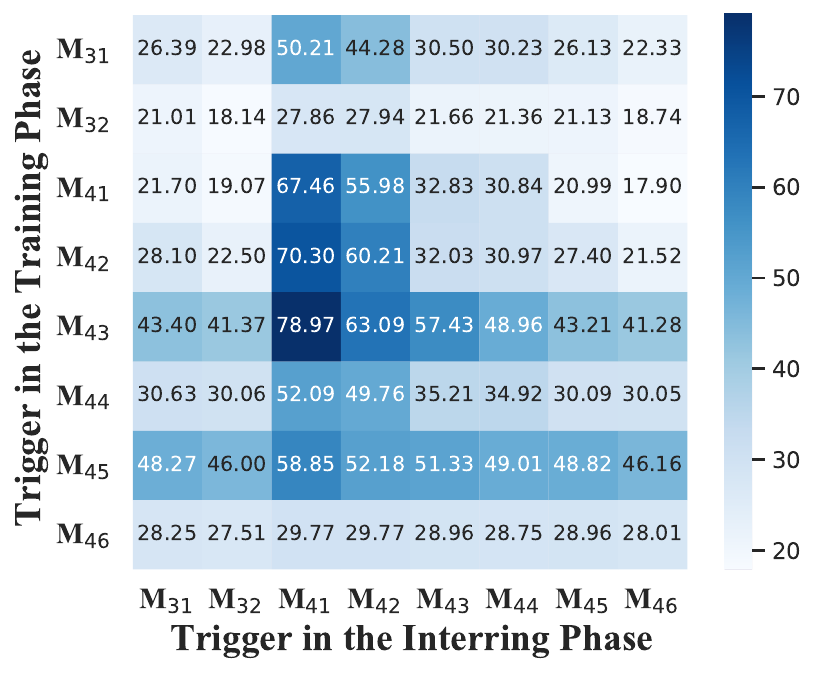}
	}%
	\centering
	\vspace{-0.1in}
	\caption{The performance of backdoor attacks launched by different motifs as the triggers on ASR in the training phase and the interring phase on the GCN model.}
	\label{fig:relitu}
\end{figure*}

\emph{3) 
	From the model and the dataset perspective, 
	the GIN model and the NIC1 dataset are more vulnerable to backdoor attacks in the three models and four datasets.}
There are differences in the attack effects of backdoor attacks on various GNNs.
For instance, 
in the AIDS dataset, 
Motif-Backdoor reaches the ASR of 88.26\%, 71.35\%, and 89.08\% on GCN, SAGPool, and GIN, respectively. 
We believe that the difference in attack effect caused by the difference in the way the model extracts network features.
Specifically,
SAGPool adopts a self-attention mask strategy for feature aggregation, 
which drops the information of some nodes.
This indicates that the trigger nodes may be dropped and affect the backdoor attack.

Furthermore, 
from the dataset point view, 
we observe that the difference in attack performance caused by the difference in the dataset is obvious. 
It is worth noting that Motif-Backdoor reaches the ASR of 99.43\% aiming at the GIN model on the NCI1 dataset, 
but only 72.78\% on the DBLP\_v1. 
The reason is that the DBLP\_v1 dataset has more graphs than other datasets, 
i.e., 
it has tens of thousands of graphs. 
In the training phase of the target model,
the number of benign graphs is much more than the number of backdoored graphs, 
which makes it more difficult for the target model to leave a backdoor.

\begin{figure*}[htbp]
	\centering
	\subfigure[ASR-GCN]{
		\includegraphics[width=0.3\linewidth, ]{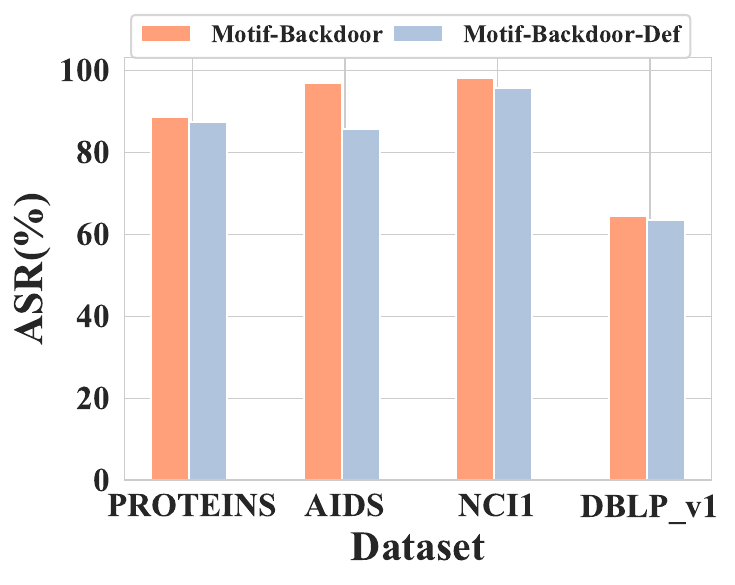}
	} 
	\subfigure[ASR-SAGPool]{
		\includegraphics[width=0.3\linewidth, ]{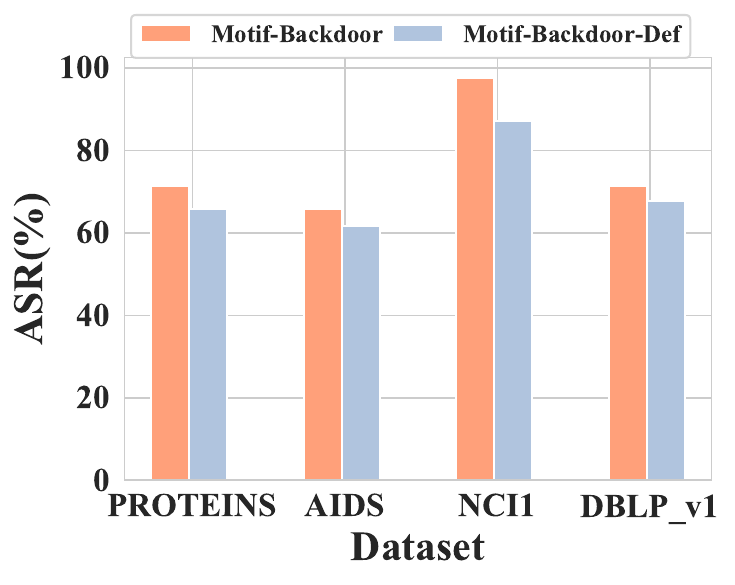}
	} 
	\subfigure[ASR-GIN]{
		\includegraphics[width=0.3\linewidth, ]{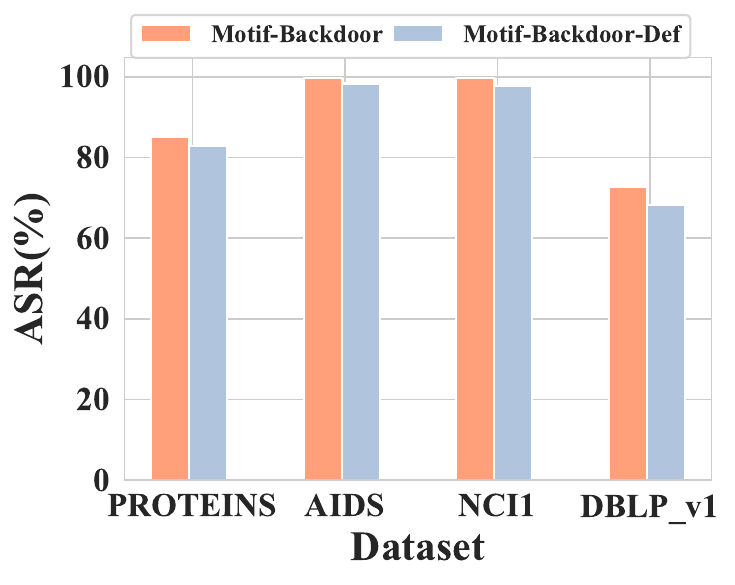}
	}%
	\\
	\subfigure[AMC-GCN]{
		\includegraphics[width=0.3\linewidth, ]{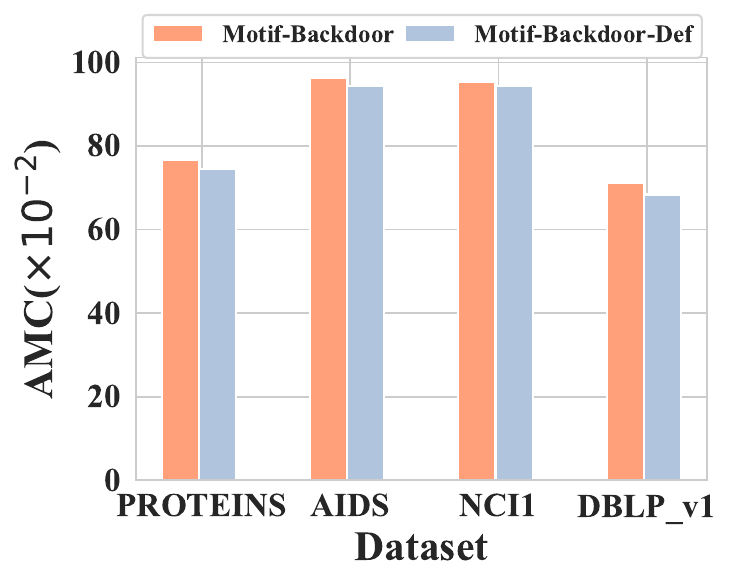}
	}
	\subfigure[AMC-SAGPool]{
		\includegraphics[width=0.3\linewidth, ]{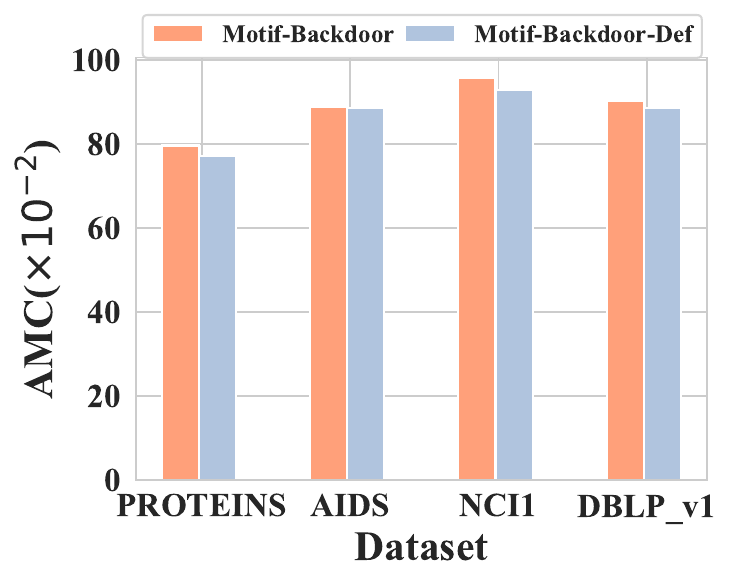}
	} 
	\subfigure[AMC-GIN]{
		\includegraphics[width=0.3\linewidth, ]{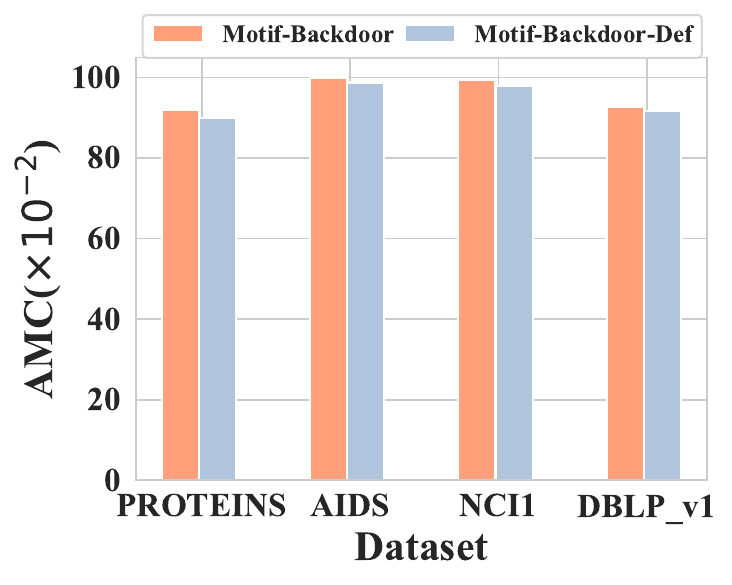}
	}%
	\centering
	\caption{ASR and AMC of the defense against Motif-Backdoor on three datasets and four models. Motif-Backdoor represents attack without defense and Motif-Backdoor-Def represents attack under defense.}
	\label{fig:def}
\end{figure*}

\subsection{A Series Triggers on the Backdoor Attack (RQ3)}
\label{Se_RQ2}
In this section,
we explore the impact of the trigger choice at different phase 
(i.e., 
the training phase and the inferring phase) 
of the backdoor attack on the attack effect. 
Specifically,
we distinguish the triggers used in the training phase and the inference phase.
According to the various motifs studied in this work, 
the backdoor attack is sequentially matched to each other on four datasets.
To more intuitively observe the change of the backdoor attack performance caused by the change of the time motif,
we directly select the motif as the trigger to randomly inject it into the benign sample to initiate the backdoor attack, 
namely Series-Backdoor. 
The results are shown in Fig.~\ref{fig:relitu}. 
Some observations are concluded in this experiment.

\emph{1) Motifs with similar structures as the trigger have similar backdoor attack effects.} 
Take the backdoor attack on the AIDS dataset as an example, 
we observed the similar attack effects from $M_{32}$ and $M_{43}$ as the trigger in the training phase or the inferring phase. 
In the case of using the trigger as $M_{32}$ in the training phase, 
$M_{32}$ and $M_{43}$ reaches the ASR of 84.13\% and 84.63\% in the interring phase. 
Intuitively, 
we can find that the color depth of the heat map brought by $M_{32}$ and $M_{43}$ is similar, 
and we group such motifs into the same series. 

The reason is that the similarity between the structures is high, 
so that the model obtains similar features when aggregating the trigger features. 
For instance,
both $M_{32}$ and $M_{43}$ are closed-loop structures. 
Intuitively, 
the model uses the mean strategy to aggregate node features. 
The extracted features based on the two motifs are the same, 
which also indicates the similarity of the two motifs. 
Therefore, 
we believe that the same series of motifs, 
i.e.,
with the similar structures, 
can replace each other in the attack.
Considering the attack effect and structure realized by the motif as a trigger,
we group $M_{31}$, $M_{41}$,  and $M_{42}$ into the same series, 
as well as $M_{44}$, $M_{45}$, and $M_{46}$.

\emph{2)
	The same motif as the trigger during the training and the inference does not show the best attack effect at the whole time.} 
It is worth noting that in most cases, 
we would consider using the same trigger for the training phase and inference phase to be the best. 
However,
in the case of using the trigger as $M_{44}$ in the training phase on the PROTEINS dataset, 
$M_{44}$ and $M_{46}$ reaches the ASR of 63.33\% and 71.75\% in the interring phase. 
Besides, 
there are several other examples with similar situations, 
e.g., 
$M_{42}$ in the training phase on the NCI1 dataset.

We speculate the reason is that 
there is a mutually inclusive relationship between the motif structures.
Specifically, 
$M_{46}$ contains the structures of the remaining motifs, 
and they can all find corresponding subgraphs in $M_{46}$. 
This means that $M_{46}$ can activate the backdoor left by the remaining motifs as triggers in the training. 
But it does not mean that the more complex the trigger, 
the better the attack. 
For instance, 
$M_{41}$ and $M_{42}$ can accomplish better attack results than $M_{46}$ in DBLP\_v1 dataset. 
In addition to considering the complexity of the trigger structure,
it is also necessary to consider the distribution of the motifs in the dataset to reach an effective backdoor attack.

\subsection{Defense against Motif-Backdoor (RQ4)}
We evaluate the attack performance of Motif-Backdoor under possible defenses. 
Due to the current lack of research on the backdoor attack defense in GNNs, 
inspired by Wu \textit{et al.}~\cite{wu2019adversarial}, 
we transfer the defense method  against the adversarial attack to evaluate the Motif-Backdoor.
They used the Jaccard index to calculate the similarity between nodes. 
Based on the node similarity, 
they filtered and removed abnormal connections, 
realizing defense against adversarial attacks.
For the backdoor attack, 
the attacker need to use the trigger to activate the backdoor in the model, 
causing the model to predict incorrectly. 
This also means that the defender can destroy the triggers in the input graphs,
thus achieving the purpose of defending against the backdoor attack. 
Specifically, 
in the inference, 
we use the Jaccard index to calculate the similarity of nodes in the graph, 
dropping the links whose similarity ranks in the bottom 10\%. 
We conduct defense experiments on three models and four datasets, 
where Motif-Backdoor-Def represents Motif-Backdoor with the Jaccard defense mechanism. 
The results are shown in Fig.~\ref{fig:def}.

We can observe that Motif-Backdoor and Motif-Backdoor-Def perform similarly on ASR and AMC. 
Take the backdoor attack on the AIDS dataset on the GIN model as an example, 
Motif-Backdoor achieves the ASR of 99.72\% and the AMC of 0.9943, 
while Motif-Backdoor-Def reaches the ASR of 97.68\% and the AMC of 0.9795. 
According to the attack results of Motif-Backdoor-Def, 
the defense mechanism based on Jaccard index cannot play a satisfactory defense effect.
The reason is that the defense method of dropping the part links may destroy the integrity of the trigger, 
but it is difficult to completely remove the trigger from the input data.
In Sec.~\ref{Se_RQ2}, 
it has been verified that the same series of subgraphs can also activate the backdoor in the model. 
We conclude that the defense method of deleting the part links leads to the fact that the trigger subgraph structure only changes in the same series, 
which can still be used for the backdoor attack.

\subsection{Parameter Sensitivity Analysis and Visualization (RQ5)}
In the section, 
we first analyze the influence of several key hyper-parameters in different ranges on the effect of Motif-Backdoor. 
Then,
we visualize the benign graphs as well as the backdoored graphs to explore the stealthiness of Motif-Backdoor intuitively.

\begin{figure}[htbp]
	\centering
	
	\subfigure[GCN]{
		\includegraphics[width=0.45\linewidth]{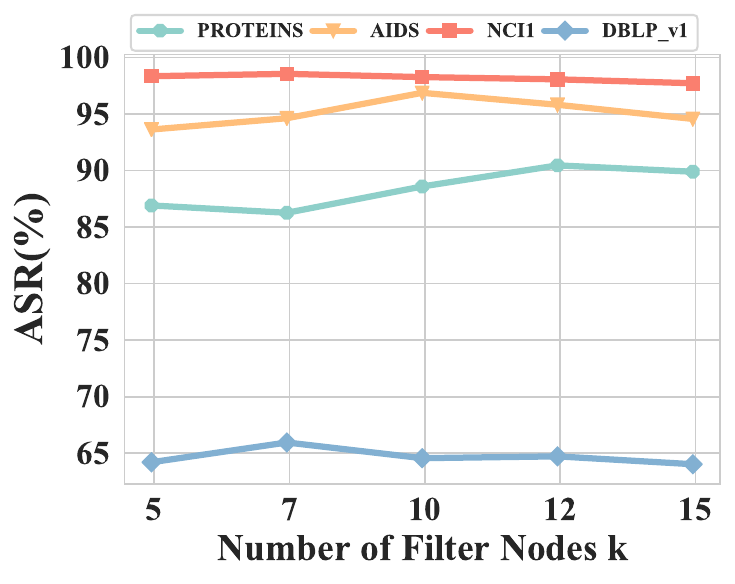}
	} 
	\subfigure[GIN]{
		\includegraphics[width=0.45\linewidth]{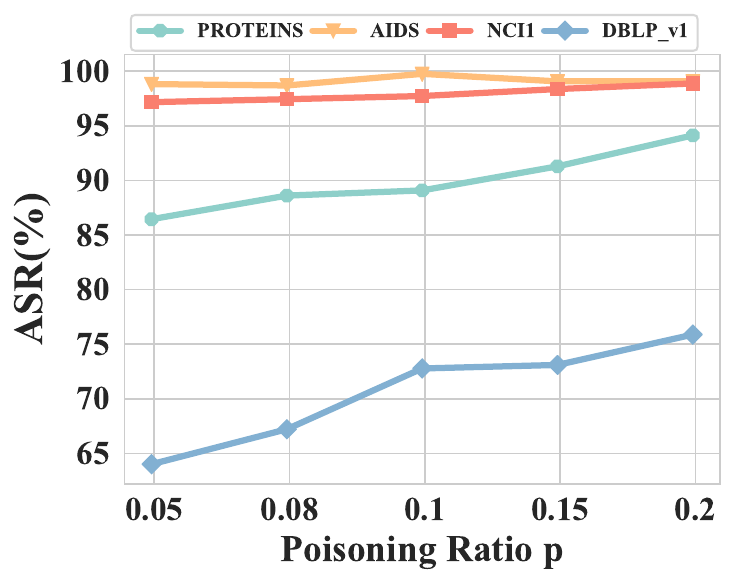}
		
	}
	\caption{Parameter sensitivity analysis of Motif-Backdoor about $k$ and $p$.}
	\label{fig:params}
\end{figure}

\textbf{Parameter Sensitivity Analysis}.
We study the sensitivity of several key hyper-parameters, 
i.e., 
number of filter node $k$ and poisoning rate $p$, 
at different scales.
Motif-Backdoor performs hyper-parameter analysis on four datasets, 
as shown in Fig.~\ref{fig:params}. 
For the number of filter nodes $k$, 
we observe that the performance of Motif-Backdoor is stable under the change of $k$. 
For instance, 
on the NCI1 dataset and the GCN model in Fig.~\ref{fig:params}(a), Motif-Backdoor reaches the ASR of 98.34\%, 98.55\%, 98.26\%, 98.06\%, and 97.72\% in the range of the number of filtered nodes, respectively. 
The possible reason is that the nodes selected by the graph importance index take into account the graph structure. 
They need further feedback from the shadow model to determine the trigger injection position. 
This also indicates that the trigger injection position is more dependent on the $subscore$. 
The graph index plays the role of filtering unimportant nodes for the first time,
which reduces the number of nodes that are subsequently considered by the model.

In Fig.~\ref{fig:params}(b), 
for the poison rate $p$, 
the ASR achieved by Motif-Backdoor increases as the poison rate increases. 
Take the backdoor attack on the DBLP\_v1 dataset on the GIN model as an example, 
Motif-Backdoor achieves the ASR of 64.01\%, 67.22\%, 72.78\%, 73.10\%, and 75.88\% in the range of the number of filtered nodes, respectively. 
We speculate that with the increase in the poison rate $p$, 
the number of backdoored graphs has increased. 
This makes the backdoored graphs participate in the training of the target model to a greater extent, 
increasing the probability of leaving a backdoor in the target model.
Additionally,
the increase in the number of backdoored graphs also increases the risk of the backdoor attack being detected. 
Therefore,
it is critical to select an appropriate poisoning ratio.
We observe that Motif-Backdoor can achieve a satisfactory attack effect when the poisoning ratio is equal to 0.1. 
In most cases,
backdoor attacks can choose to set the poisoning ratio to 0.1, 
which takes into account the effectiveness and stealthiness of the attack.

\begin{figure}[ht]
	\centering
	\subfigure[PROTEINS]{
		\includegraphics[width=0.45\linewidth]{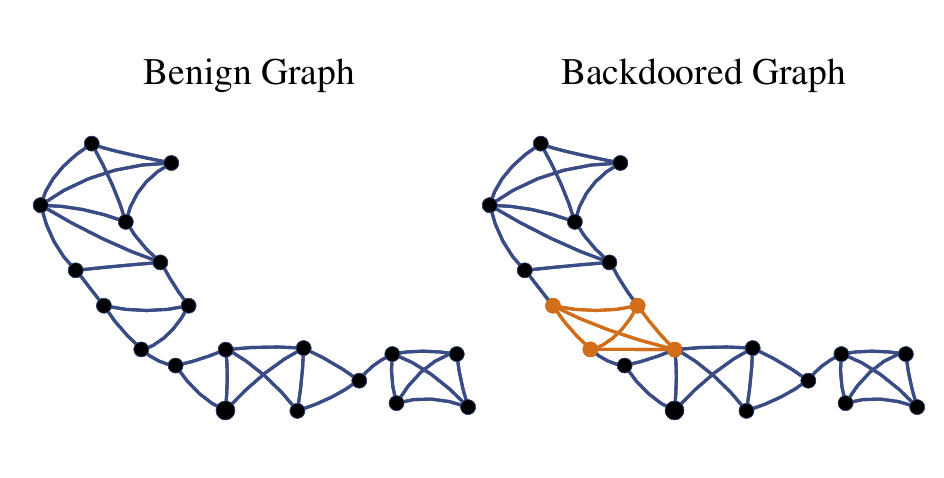}
	}
	\subfigure[AIDS]{
		\includegraphics[width=0.45\linewidth]{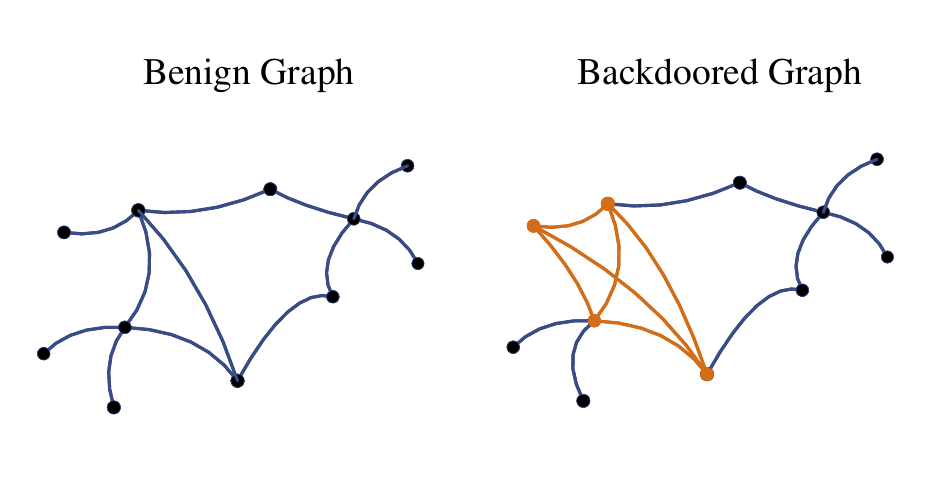}
	}
	\\
	\subfigure[NCI1]{
		\includegraphics[width=0.45\linewidth]{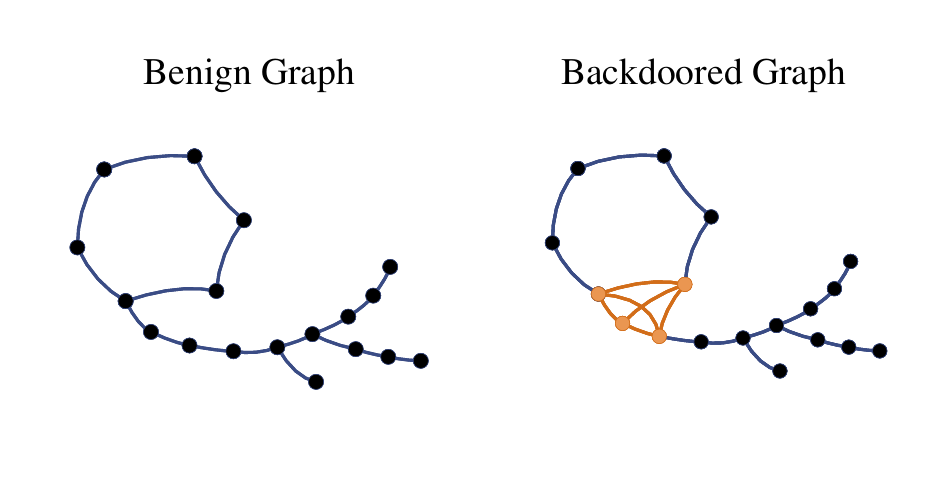}
	}
	\subfigure[DBLP\_v1]{
		\includegraphics[width=0.45\linewidth]{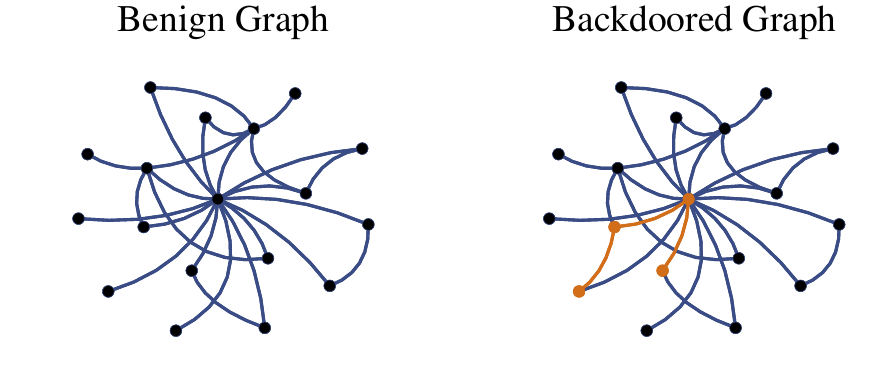}
	}
	\caption{Visualization of the benign graphs and the backdoored garphs on the four datasets. Blue links: the target link to be predicted. Orange links:  the structure of the trigger by the Motif-Backdoor.}
	\label{fig:vis}
\end{figure}

\textbf{Visualization.} 
To explore the stealthiness of Motif-Backdoor, 
we visualize the backdoored graphs by the Motif-Backdoor on three models and four datasets,
as shown in Fig.~\ref{fig:vis}. 
Intuitively, 
Motif-Backdoor does not do much damage to the benign graphs,
and only a few links are needed to modify the graph to conduce effective attack.
The visualization of the benign graphs and the backdoored graphs is similar. 
We speculate that the trigger injection position plays a key role in the stealthiness of the graph. 
Motif-Backdoor considers the graph structure and model feedback to select an appropriate trigger injection position.

\section{Conclusion\label{Cons}}
In this work, 
based on the differences in the distribution of network motifs, 
we obtain two interesting insights on the backdoor attack. 
Then,
we propose a base motif backdoor attack framework against GNNs, 
named Motif-Backdoor
which uses the distribution difference of the motifs in the dataset to search the trigger structure. 
Besides, 
it adopts the graph importance index and the defined subgraph score to find the trigger injection position.
Extensive experiments show Motif-Backdoor can achieve the SOTA performance. 
Even with the deployment of the defense mechanism, 
Motif-Backdoor still realizes impressive attack results.

However, 
Motif-Backdoor is still challenged in several aspects.
It requires the confidence score of the input sample by accessing the target model extensively to construct the shadow model,
which increases the risk of an attack being detected. 
Moreover,
it is necessary to further pay attention to effective defense strategies on GNNs in different downstream tasks, 
e.g., node classification, 
edge classification and 
structural role classification. 
The interpretability of backdoor attacks is also interesting.

	\bibliographystyle{IEEEtran}      
	\bibliography{Reference}                        

\begin{thebibliography}{10}
\providecommand{\url}[1]{#1}
\csname url@samestyle\endcsname
\providecommand{\newblock}{\relax}
\providecommand{\bibinfo}[2]{#2}
\providecommand{\BIBentrySTDinterwordspacing}{\spaceskip=0pt\relax}
\providecommand{\BIBentryALTinterwordstretchfactor}{4}
\providecommand{\BIBentryALTinterwordspacing}{\spaceskip=\fontdimen2\font plus
\BIBentryALTinterwordstretchfactor\fontdimen3\font minus
  \fontdimen4\font\relax}
\providecommand{\BIBforeignlanguage}[2]{{%
\expandafter\ifx\csname l@#1\endcsname\relax
\typeout{** WARNING: IEEEtran.bst: No hyphenation pattern has been}%
\typeout{** loaded for the language `#1'. Using the pattern for}%
\typeout{** the default language instead.}%
\else
\language=\csname l@#1\endcsname
\fi
#2}}
\providecommand{\BIBdecl}{\relax}
\BIBdecl

\bibitem{zhang2022influence}
X.-J. Zhang, J.~Wang, X.-J. Ma, C.~Ma, J.-Q. Kan, and H.-F. Zhang, ``Influence
  maximization in social networks with privacy protection,'' \emph{Physica A:
  Statistical Mechanics and its Applications}, vol. 607, p. 128179, 2022.

\bibitem{zola2022network}
F.~Zola, L.~Segurola-Gil, J.~L. Bruse, M.~Galar, and R.~Orduna-Urrutia,
  ``Network traffic analysis through node behaviour classification: a
  graph-based approach with temporal dissection and data-level preprocessing,''
  \emph{Computers \& Security}, vol. 115, p. 102632, 2022.

\bibitem{chen2020software}
J.~Chen, K.~Hu, Y.~Yu, Z.~Chen, Q.~Xuan, Y.~Liu, and V.~Filkov, ``Software
  visualization and deep transfer learning for effective software defect
  prediction,'' in \emph{Proceedings of the ACM/IEEE 42nd international
  conference on software engineering}, 2020, pp. 578--589.

\bibitem{Hong22Mul}
H.~Yang, H.~He, W.~Zhang, and Y.~Bai, ``{MTGK:} multi-source cross-network node
  classification via transferable graph knowledge,'' \emph{Inf. Sci.}, vol.
  589, pp. 395--415, 2022.

\bibitem{Wen22Hi}
W.~Yao, K.~Guo, Y.~Hou, and X.~Li, ``Hierarchical structure-feature aware graph
  neural network for node classification,'' \emph{{IEEE} Access}, vol.~10, pp.
  36\,846--36\,855, 2022.

\bibitem{Zh18Sa}
Z.~Ying, J.~You, C.~Morris, X.~Ren, W.~L. Hamilton, and J.~Leskovec,
  ``Hierarchical graph representation learning with differentiable pooling,''
  in \emph{Advances in Neural Information Processing Systems 31: Annual
  Conference on Neural Information Processing Systems 2018, NeurIPS 2018,
  December 3-8, 2018, Montr{\'{e}}al, Canada}, 2018, pp. 4805--4815.

\bibitem{Ke19How}
K.~Xu, W.~Hu, J.~Leskovec, and S.~Jegelka, ``How powerful are graph neural
  networks?'' in \emph{7th International Conference on Learning
  Representations, {ICLR} 2019, New Orleans, LA, USA, May 6-9, 2019}.\hskip 1em
  plus 0.5em minus 0.4em\relax OpenReview.net, 2019, pp. 1--17.

\bibitem{xia2022machine}
X.~Xia, Y.~Su, L.~L{\"u}, X.~Zhang, Y.-C. Lai, and H.-F. Zhang, ``Machine
  learning prediction of network dynamics with privacy protection,''
  \emph{Physical Review Research}, vol.~4, no.~4, p. 043076, 2022.

\bibitem{Mu18Link}
M.~Zhang and Y.~Chen, ``Link prediction based on graph neural networks,'' in
  \emph{Advances in Neural Information Processing Systems 31: Annual Conference
  on Neural Information Processing Systems 2018, NeurIPS 2018, December 3-8,
  2018, Montr{\'{e}}al, Canada}, 2018, pp. 5171--5181.

\bibitem{zheng2022neuronfair}
H.~Zheng, Z.~Chen, T.~Du, X.~Zhang, Y.~Cheng, S.~Ji, J.~Wang, Y.~Yu, and
  J.~Chen, ``Neuronfair: Interpretable white-box fairness testing through
  biased neuron identification,'' in \emph{Proceedings of the 44th
  International Conference on Software Engineering}, 2022, pp. 1519--1531.

\bibitem{jia2023topology}
M.~Jia, J.~Hu, Y.~Liu, Z.~Gao, and Y.~Yao, ``Topology-guided graph learning for
  process fault diagnosis,'' \emph{Industrial \& Engineering Chemistry
  Research}, vol.~62, no.~7, pp. 3238--3248, 2023.

\bibitem{liu2022deep}
K.~Liu, M.~Zheng, Y.~Liu, J.~Yang, and Y.~Yao, ``Deep autoencoder thermography
  for defect detection of carbon fiber composites,'' \emph{IEEE Transactions on
  Industrial Informatics}, pp. 1--10, 2022.

\bibitem{zheng2021grip}
H.~Zheng, J.~Chen, H.~Du, W.~Zhu, S.~Ji, and X.~Zhang, ``Grip-gan: An
  attack-free defense through general robust inverse perturbation,'' \emph{IEEE
  Transactions on Dependable and Secure Computing}, vol.~19, no.~6, pp.
  4204--4224, 2021.

\bibitem{jiang2022holistic}
B.~Jiang, Y.~Liu, H.~Geng, Y.~Wang, H.~Zeng, and J.~Ding, ``A holistic feature
  selection method for enhanced short-term load forecasting of power system,''
  \emph{IEEE Transactions on Instrumentation and Measurement}, pp. 1--11, 2022.

\bibitem{chen2023egc2}
J.~Chen, H.~Xiong, H.~Zheng, D.~Zhang, J.~Zhang, M.~Jia, and Y.~Liu, ``Egc2:
  Enhanced graph classification with easy graph compression,''
  \emph{Information Sciences}, vol. 629, pp. 376--397, 2023.

\bibitem{Jun19Ka}
J.~Lee, I.~Lee, and J.~Kang, ``Self-attention graph pooling,'' in
  \emph{Proceedings of the 36th International Conference on Machine Learning,
  {ICML} 2019, 9-15 June 2019, Long Beach, California, {USA}}, ser. Proceedings
  of Machine Learning Research, vol.~97.\hskip 1em plus 0.5em minus 0.4em\relax
  {PMLR}, 2019, pp. 3734--3743.

\bibitem{zhuang2020smart}
Y.~Zhuang, Z.~Liu, P.~Qian, Q.~Liu, X.~Wang, and Q.~He, ``Smart contract
  vulnerability detection using graph neural network.'' in \emph{IJCAI}, 2020,
  pp. 3283--3290.

\bibitem{hu2022smpc}
W.~Hu, X.~Xia, X.~Ding, X.~Zhang, K.~Zhong, and H.-F. Zhang, ``Smpc-ranking: A
  privacy-preserving method on identifying influential nodes in multiple
  private networks,'' \emph{IEEE Transactions on Systems, Man, and Cybernetics:
  Systems}, pp. 1--12, 2022.

\bibitem{jia2023graph}
M.~Jia, D.~Xu, T.~Yang, Y.~Liu, and Y.~Yao, ``Graph convolutional network soft
  sensor for process quality prediction,'' \emph{Journal of Process Control},
  vol. 123, pp. 12--25, 2023.

\bibitem{Br14So}
B.~Perozzi, R.~Al{-}Rfou, and S.~Skiena, ``Deepwalk: online learning of social
  representations,'' in \emph{The 20th {ACM} {SIGKDD} International Conference
  on Knowledge Discovery and Data Mining, {KDD} '14, New York, NY, {USA} -
  August 24 - 27, 2014}.\hskip 1em plus 0.5em minus 0.4em\relax {ACM}, 2014,
  pp. 701--710.

\bibitem{Ad16Ba}
A.~Grover and J.~Leskovec, ``node2vec: Scalable feature learning for
  networks,'' in \emph{Proceedings of the 22nd {ACM} {SIGKDD} International
  Conference on Knowledge Discovery and Data Mining, San Francisco, CA, USA,
  August 13-17, 2016}.\hskip 1em plus 0.5em minus 0.4em\relax {ACM}, 2016, pp.
  855--864.

\bibitem{Da18Ad}
D.~Z{\"{u}}gner, A.~Akbarnejad, and S.~G{\"{u}}nnemann, ``Adversarial attacks
  on neural networks for graph data,'' in \emph{Proceedings of the 24th {ACM}
  {SIGKDD} International Conference on Knowledge Discovery {\&} Data Mining,
  {KDD} 2018, London, UK, August 19-23, 2018}, Y.~Guo and F.~Farooq, Eds.\hskip
  1em plus 0.5em minus 0.4em\relax {ACM}, 2018, pp. 2847--2856.

\bibitem{Yao21Gra}
Y.~Ma, S.~Wang, T.~Derr, L.~Wu, and J.~Tang, ``Graph adversarial attack via
  rewiring,'' in \emph{{KDD} '21: The 27th {ACM} {SIGKDD} Conference on
  Knowledge Discovery and Data Mining, Virtual Event, Singapore, August 14-18,
  2021}.\hskip 1em plus 0.5em minus 0.4em\relax {ACM}, 2021, pp. 1161--1169.

\bibitem{Jin20Ad}
J.~Li, T.~Xie, L.~Chen, F.~Xie, X.~He, and Z.~Zheng, ``Adversarial attack on
  large scale graph,'' \emph{IEEE Transactions on Knowledge and Data
  Engineering}, vol.~35, no.~1, pp. 82--95, 2021.

\bibitem{chen2021time}
J.~Chen, J.~Zhang, Z.~Chen, M.~Du, and Q.~Xuan, ``Time-aware gradient attack on
  dynamic network link prediction,'' \emph{IEEE Transactions on Knowledge \&
  Data Engineering}, vol.~35, no.~02, pp. 2091--2102, 2023.

\bibitem{Za21Back}
Z.~Zhang, J.~Jia, B.~Wang, and N.~Z. Gong, ``Backdoor attacks to graph neural
  networks,'' in \emph{{SACMAT} '21: The 26th {ACM} Symposium on Access Control
  Models and Technologies, Virtual Event, Spain, June 16-18, 2021}.\hskip 1em
  plus 0.5em minus 0.4em\relax {ACM}, 2021, pp. 15--26.

\bibitem{Jing21Exp}
J.~Xu, M.~Xue, and S.~Picek, ``Explainability-based backdoor attacks against
  graph neural networks,'' in \emph{WiseML@WiSec 2021: Proceedings of the 3rd
  {ACM} Workshop on Wireless Security and Machine Learning, Abu Dhabi, United
  Arab Emirates, July 2, 2021}.\hskip 1em plus 0.5em minus 0.4em\relax {ACM},
  2021, pp. 31--36.

\bibitem{Zhao21Graph}
Z.~Xi, R.~Pang, S.~Ji, and T.~Wang, ``Graph backdoor,'' in \emph{30th {USENIX}
  Security Symposium, {USENIX} Security 2021, August 11-13, 2021}, M.~Bailey
  and R.~Greenstadt, Eds.\hskip 1em plus 0.5em minus 0.4em\relax {USENIX}
  Association, 2021, pp. 1523--1540.

\bibitem{Yu21Back}
Y.~Sheng, R.~Chen, G.~Cai, and L.~Kuang, ``Backdoor attack of graph neural
  networks based on subgraph trigger,'' in \emph{Collaborative Computing:
  Networking, Applications and Worksharing - 17th {EAI} International
  Conference, CollaborateCom 2021, Virtual Event, October 16-18, 2021,
  Proceedings, Part {II}}, ser. Lecture Notes of the Institute for Computer
  Sciences, Social Informatics and Telecommunications Engineering, vol.
  407.\hskip 1em plus 0.5em minus 0.4em\relax Springer, 2021, pp. 276--296.

\bibitem{milo2002network}
R.~Milo, S.~Shen-Orr, S.~Itzkovitz, N.~Kashtan, D.~Chklovskii, and U.~Alon,
  ``Network motifs: simple building blocks of complex networks,''
  \emph{Science}, vol. 298, no. 5594, pp. 824--827, 2002.

\bibitem{alon2007network}
U.~Alon, ``Network motifs: theory and experimental approaches,'' \emph{Nature
  Reviews Genetics}, vol.~8, no.~6, pp. 450--461, 2007.

\bibitem{yin2017local}
H.~Yin, A.~R. Benson, J.~Leskovec, and D.~F. Gleich, ``Local higher-order graph
  clustering,'' in \emph{Proceedings of the 23rd ACM SIGKDD international
  conference on knowledge discovery and data mining}, 2017, pp. 555--564.

\bibitem{sankar2017motif}
A.~Sankar, X.~Zhang, and K.~C.-C. Chang, ``Motif-based convolutional neural
  network on graphs,'' \emph{arXiv preprint arXiv:1711.05697}, 2017.

\bibitem{yang2018node}
C.~Yang, M.~Liu, V.~W. Zheng, and J.~Han, ``Node, motif and subgraph:
  Leveraging network functional blocks through structural convolution,'' in
  \emph{2018 IEEE/ACM International Conference on Advances in Social Networks
  Analysis and Mining (ASONAM)}.\hskip 1em plus 0.5em minus 0.4em\relax IEEE,
  2018, pp. 47--52.

\bibitem{zhao2019motif}
H.~Zhao, Y.~Zhou, Y.~Song, and D.~L. Lee, ``Motif enhanced recommendation over
  heterogeneous information network,'' in \emph{Proceedings of the 28th ACM
  international conference on information and knowledge management}, 2019, pp.
  2189--2192.

\bibitem{lee2019graph}
J.~B. Lee, R.~A. Rossi, X.~Kong, S.~Kim, E.~Koh, and A.~Rao, ``Graph
  convolutional networks with motif-based attention,'' in \emph{Proceedings of
  the 28th ACM international conference on information and knowledge
  management}, 2019, pp. 499--508.

\bibitem{dareddy2019motif2vec}
M.~R. Dareddy, M.~Das, and H.~Yang, ``motif2vec: Motif aware node
  representation learning for heterogeneous networks,'' in \emph{2019 IEEE
  International Conference on Big Data (Big Data)}.\hskip 1em plus 0.5em minus
  0.4em\relax IEEE, 2019, pp. 1052--1059.

\bibitem{shao2021network}
P.~Shao, Y.~Yang, S.~Xu, and C.~Wang, ``Network embedding via motifs,''
  \emph{ACM Transactions on Knowledge Discovery from Data (TKDD)}, vol.~16,
  no.~3, pp. 1--20, 2021.

\bibitem{zhao2022motif}
M.~Zhao, Y.~Zhang, X.~Xia, and X.~Xu, ``Motif-aware adversarial graph
  representation learning,'' \emph{IEEE Access}, vol.~10, pp. 8617--8626, 2022.

\bibitem{wang2020model}
L.~Wang, J.~Ren, B.~Xu, J.~Li, W.~Luo, and F.~Xia, ``Model: Motif-based deep
  feature learning for link prediction,'' \emph{IEEE Transactions on
  Computational Social Systems}, vol.~7, no.~2, pp. 503--516, 2020.

\bibitem{yu2022motifexplainer}
Z.~Yu and H.~Gao, ``Motifexplainer: a motif-based graph neural network
  explainer,'' \emph{arXiv preprint arXiv:2202.00519}, 2022.

\bibitem{Pre21Jing}
J.~Piao, G.~Zhang, F.~Xu, Z.~Chen, and Y.~Li, ``Predicting customer value with
  social relationships via motif-based graph attention networks,'' in
  \emph{{WWW} '21: The Web Conference 2021, Virtual Event / Ljubljana,
  Slovenia, April 19-23, 2021}.\hskip 1em plus 0.5em minus 0.4em\relax {ACM} /
  {IW3C2}, 2021, pp. 3146--3157.

\bibitem{hovcevar2014combinatorial}
T.~Ho{\v{c}}evar and J.~Dem{\v{s}}ar, ``A combinatorial approach to graphlet
  counting,'' \emph{Bioinformatics}, vol.~30, no.~4, pp. 559--565, 2014.

\bibitem{borgwardt2005protein}
K.~M. Borgwardt, C.~S. Ong, S.~Sch{\"{o}}nauer, S.~V.~N. Vishwanathan, A.~J.
  Smola, and H.~Kriegel, ``Protein function prediction via graph kernels,'' in
  \emph{Proceedings Thirteenth International Conference on Intelligent Systems
  for Molecular Biology 2005, Detroit, MI, USA}, 2005, pp. 47--56.

\bibitem{rossi2015network}
R.~Rossi and N.~Ahmed, ``The network data repository with interactive graph
  analytics and visualization,'' in \emph{Proceedings of the AAAI conference on
  artificial intelligence}, vol.~29, no.~1, 2015, pp. 1--2.

\bibitem{shervashidze2011weisfeiler}
N.~Shervashidze, P.~Schweitzer, E.~J. van Leeuwen, K.~Mehlhorn, and K.~M.
  Borgwardt, ``Weisfeiler-lehman graph kernels,'' \emph{Journal of Machine
  Learning Research}, vol.~12, pp. 2539--2561, 2011.

\bibitem{pan2013graph}
S.~Pan, X.~Zhu, C.~Zhang, and S.~Y. Philip, ``Graph stream classification using
  labeled and unlabeled graphs,'' in \emph{2013 IEEE 29th International
  Conference on Data Engineering (ICDE)}.\hskip 1em plus 0.5em minus
  0.4em\relax IEEE, 2013, pp. 398--409.

\bibitem{freeman1978centrality}
L.~C. Freeman, ``Centrality in social networks conceptual clarification,''
  \emph{Social networks}, vol.~1, no.~3, pp. 215--239, 1978.

\bibitem{lee2019self}
J.~Lee, I.~Lee, and J.~Kang, ``Self-attention graph pooling,'' in
  \emph{International conference on machine learning}.\hskip 1em plus 0.5em
  minus 0.4em\relax PMLR, 2019, pp. 3734--3743.

\bibitem{Th17Se}
T.~N. Kipf and M.~Welling, ``Semi-supervised classification with graph
  convolutional networks,'' in \emph{5th International Conference on Learning
  Representations, {ICLR} 2017, Toulon, France, April 24-26, 2017, Conference
  Track Proceedings}.\hskip 1em plus 0.5em minus 0.4em\relax OpenReview.net,
  2017, pp. 1--14.

\bibitem{chen2021graphattacker}
J.~Chen, D.~Zhang, Z.~Ming, K.~Huang, W.~Jiang, and C.~Cui, ``Graphattacker: A
  general multi-task graph attack framework,'' \emph{IEEE Transactions on
  Network Science and Engineering}, vol.~9, no.~2, pp. 577--595, 2021.

\bibitem{wu2019adversarial}
H.~Wu, C.~Wang, Y.~Tyshetskiy, A.~Docherty, K.~Lu, and L.~Zhu, ``Adversarial
  examples on graph data: Deep insights into attack and defense,'' \emph{arXiv
  preprint arXiv:1903.01610}, 2019.

\end{thebibliography}

\end{document}